\documentclass{article}

\PassOptionsToPackage{numbers,compress}{natbib}
\usepackage[main,final]{neurips_2026}

\usepackage[utf8]{inputenc}
\usepackage[T1]{fontenc}
\usepackage{hyperref}
\usepackage{url}
\usepackage{booktabs}
\usepackage{amsfonts}
\usepackage{nicefrac}
\usepackage{microtype}
\usepackage{xcolor}
\usepackage{graphicx}
\usepackage{subcaption}
\usepackage{longtable}
\usepackage{array}
\usepackage{rotating}
\usepackage{xspace}
\usepackage{float}
\usepackage{flafter}
\usepackage{placeins}
\usepackage{needspace}
\usepackage{algorithm}
\usepackage{algpseudocode}
\usepackage{amssymb}
\usepackage{newunicodechar}
\usepackage{siunitx}
\usepackage{etoolbox}

\usepackage{amsmath,amsfonts,bm}

\def\eqref#1{equation~\ref{#1}}

\def\1{\bm{1}}

\DeclareMathAlphabet{\mathsfit}{\encodingdefault}{\sfdefault}{m}{sl}
\SetMathAlphabet{\mathsfit}{bold}{\encodingdefault}{\sfdefault}{bx}{n}

\definecolor{bestcolor}{HTML}{D62728} 
\definecolor{secondbestcolor}{HTML}{1F77B4} 
\newcommand{\best}[1]{\textcolor{bestcolor}{\bfseries #1}}
\newcommand{\secondbest}[1]{\textcolor{secondbestcolor}{\bfseries #1}}
\robustify{\best}
\robustify{\secondbest}

\newunicodechar{▷}{\ensuremath{\triangleright}}

\title{CSO: Refining Robotic Policies via Skill Distribution Alignment and Skill-Grained Optimization}

\author{%
\textbf{Zhiyuan Xiang\textsuperscript{1} \quad Xiang Deng\textsuperscript{1}\thanks{Corresponding author: \href{mailto:dengxiang@hit.edu.cn}{\texttt{dengxiang@hit.edu.cn}}.} \quad Wanjie Tao\textsuperscript{2}} \\
\textbf{Xu Chen\textsuperscript{1} \quad Zhao Jielun\textsuperscript{3} \quad Weili Guan\textsuperscript{1}} \\[3pt]
\textsuperscript{1}Harbin Institute of Technology (Shenzhen) \\
\textsuperscript{2}Mashang Consumer Finance Co., Ltd. \\
\textsuperscript{3}Shenzhen Institute for Advanced Study, \\
University of Electronic Science and Technology of China
}

\hypersetup{
    colorlinks=true,
    linkcolor=blue,
    filecolor=magenta,
    urlcolor=teal,
    citecolor=blue,
    pdftitle={CSO: Refining Robotic Policies via Skill Distribution Alignment and Skill-Grained Optimization},
    pdfauthor={Zhiyuan Xiang, Xiang Deng, Wanjie Tao, Xu Chen, Zhao Jielun, Weili Guan},
}

\newcommand{\modelname}{Cascaded Skills Optimization\xspace}
\newcommand{\modelacr}{CSO\xspace}

\newcommand{\rsft}{Rejection-Sampling Supervised Fine-tuning\xspace}
\newcommand{\spo}{Skills Policy Optimization\xspace}
\newcommand{\rsftshort}{R-SFT\xspace}
\newcommand{\sposhort}{SPO\xspace}

\begin{document}

\maketitle

\begin{abstract}
Discretizing continuous actions into discrete skills using methods like VQ-VAE has emerged as a powerful paradigm for robotic manipulation. However, quantization errors yield a suboptimal distribution for the skill prior, degrading its performance. While reinforcement learning offers a path for refinement, its direct application faces unstable encoder updates and a granularity dilemma in importance sampling. To address these challenges, we introduce \modelname (\modelacr), a two-stage post-training framework. First, \rsft aligns the model's observation-to-skill mapping with the distribution of successful online trajectories via supervised fine-tuning. Second, with the observation encoder frozen, \spo computes an independent, clipped importance ratio for each complete skill, enabling stable and efficient updates. With a lightweight QueST-style policy, CSO raises overall LIBERO success from 80.1\% to 97.5\%, improves performance on RoboTwin2.0, and achieves 61.0\% success across five physical single-arm tasks. Additional comparisons with FAST, discrete BEAST, and fixed-buffer IQL further validate the effectiveness of our two-stage post-training strategy.

\end{abstract}

\section{Introduction}
\label{sec:introduction}

Modeling complex manipulation behaviors has long been a central problem in robotics~\citep{levine2016endtoend, visualforesight}. A promising direction is to learn structured action representations through Transformer-based architectures~\citep{rt12022arxiv, octo}. Temporal action chunks provide useful abstractions for control~\citep{act}; skill-based approaches additionally map continuous trajectories into discrete latent representations~\citep{vq-bet}. A quantized autoencoder, such as a VQ-VAE~\citep{vqvae, rethinking}, learns a vocabulary of fundamental tokens. These tokens are then used to form "skills", which are reusable motion primitives represented as token sequences~\citep{mete2024quest, li2025star}. This hierarchical framework simplifies the problem, enabling an autoregressive policy to tackle complex tasks by sequentially composing tokens to generate a skill~\citep{mete2024quest}.

However, while this discretization paradigm is powerful, policies trained purely via imitation learning (IL) with these discrete skills exhibit a notable limitation. The VQ-VAE's discretization process introduces inherent quantization errors, meaning the skills are only approximations of the expert's continuous actions~\citep{vq-bet}. These reconstruction errors compound with policy errors during online execution. On one hand, the inherent imprecision of the tokens that constitute a skill can cause the decoded action to fail the task. On the other hand, these quantization errors can accumulate over an episode, progressively pushing the agent into states outside the expert data distribution. This resulting distribution shift severely hinders the generalization of the IL-trained policy~\citep{kim2025fine}.

To overcome the limitations of the initial policy, a natural remedy is to refine it with online reinforcement learning (RL)~\citep{guo2025improving, liu2025can, tan2025interactive}. However, directly applying RL to this setting presents several challenges. A primary issue is the unstable update of the observation encoder. Online data collection often yields an imbalanced mix of task outcomes (e.g., many successes on easy tasks, many failures on hard ones). This provides inconsistent gradient signals to the crucial observation encoder, leading to inefficient training and often requiring complex hybrid frameworks to stabilize~\citep{guo2025improving, chen2025conrft}. A second structural challenge arises from a mismatch between the unit of decision-making (a skill) and the unit of optimization (a token). Token-level methods apply a separate importance ratio to each generated token~\citep{schulman2017proximal, lu2025vla}, which treats the components of a jointly executed skill separately. Episode-level methods such as GSPO instead use one length-normalized ratio and a shared clipping decision for the complete sequence~\citep{zheng2025groupsequencepolicyoptimization}. Robotic episodes exhibit \textit{length asymmetry}: time-outs are often longer than successes. This motivates skill-level optimization, which assigns each complete skill its own ratio and clipping decision while sharing the terminal advantage. We compare token-, skill-, and episode-level optimization from the same R-SFT checkpoint.

To address these key challenges, we propose \textbf{\modelname} (\textbf{\modelacr}), a two-stage post-training framework. Our key insight is that effective refinement requires two steps: first, stabilizing the model on successful online data, and second, aligning the unit of optimization with the unit of decision-making. Specifically:
1) To address distributional shift and encoder instability, we introduce \textbf{\rsft} (\textbf{\rsftshort}) for skill-based policy refinement. This stage aligns the model's observation-to-skill mapping with the distribution of successful online trajectories via supervised fine-tuning, inspired by similar techniques in LLMs~\citep{yuan2023scalingrelationshiplearningmathematical}. The encoder is then stabilized and frozen, providing a robust foundation for the subsequent RL stage and avoiding the inconsistent gradients of direct online fine-tuning.
2) To solve the granularity mismatch, we introduce the \textbf{\spo} (\textbf{\sposhort}) algorithm. By computing and clipping an importance ratio for each complete skill, SPO aligns optimization with execution and enables stable, data-efficient policy improvement.

We validate our framework through extensive experiments on challenging manipulation benchmarks. In its lightweight instantiation, \modelacr demonstrates substantial improvements over its underlying imitation-learned policy, boosting the average success rate by 17.4 absolute percentage points across all four LIBERO suites~\citep{libero}. We further validate the method on RoboTwin2.0~\citep{chen2025robotwin2} and confirm its real-world applicability by successfully deploying the learned policy on a physical robot.

To summarize, our main contributions are:
\vspace{-0.5em}
\begin{enumerate}
    \item To circumvent unstable encoder updates in direct reinforcement learning, we introduce \textbf{\rsftshort}. This rejection-sampling fine-tuning stage first stabilizes the observation encoder on successful online trajectories to establish a robust foundation for policy optimization.
    \item To resolve the granularity dilemma between internal tokens and complete trajectories, we propose \textbf{\sposhort}, a novel algorithm that performs stable and fine-grained policy optimization via skill-level importance sampling.
    \item We design and validate \textbf{\modelacr}, a complete two-stage framework that systematically refines skill-based policies by decoupling representation stabilization from policy optimization, demonstrating substantial performance improvements on challenging benchmarks and real-world robotic tasks.
\end{enumerate}

\section{Related Work}
\label{sec:related_work}

\subsection{Action Representations in Imitation Learning}
Action chunking, as in ACT~\citep{act}, represents temporally extended control, while VQ-BeT~\citep{vq-bet}, QueST~\citep{mete2024quest}, and STAR~\citep{li2025star} learn discrete action representations using quantized autoencoders~\citep{vqvae, rethinking}. Generalist policies such as RT-2~\citep{rt2}, Octo~\citep{octo}, and OpenVLA~\citep{kim2024openvla} further connect perception and language to control. F1~\citep{lv2025f1} uses generated visual foresight to guide action prediction. Alternative representations include FAST's frequency-based compression~\citep{fast}, BEAST's B-splines~\citep{zhou2025beast}, and Hermite trajectory priors for smoothness and inter-chunk continuity~\citep{lv2026hermite}. Reconstruction quality and closed-loop execution provide complementary criteria for evaluating action representations. Our matched comparison shows that R-SFT and full CSO improve policy success with FAST and discrete BEAST (Section~\ref{sec:controlled}). We evaluate FAST and discrete BEAST through a shared autoregressive policy interface.

\subsection{Reinforcement Learning for Large Language Models}

Work in LLMs shows that RL after pre-training and SFT can unlock stronger reasoning and interactive adaptation~\citep{ouyang2022training, guo2025deepseek, shao2024deepseekmath, liu2024direct, liu2025understanding}. Beyond PPO~\citep{schulman2017proximal} and DPO~\citep{rafailov2023direct}, value-free methods such as Rejection Sampling Fine-Tuning, GRPO, and GSPO have become increasingly influential~\citep{yuan2023scalingrelationshiplearningmathematical, shao2024deepseekmath, zheng2025groupsequencepolicyoptimization}. Our setting differs because sequence-level optimization creates a granularity mismatch in robotics, which SPO resolves with skill-level updates.

\subsection{Reinforcement Learning for Robotic Policies}

Recent work increasingly applies RL to refine pretrained robotic policies~\citep{guo2025improving, liu2025can}. Prior approaches adapt PPO-like algorithms~\citep{lu2025vla}, dense rewards~\citep{zhang2025reinbot}, learned critics~\citep{shu2025rftf}, or preference-style optimization~\citep{zhang2024grape, chen2025tgrpo}, often with substantial auxiliary machinery~\citep{guo2025improving, chen2025conrft}. Concurrent work such as RIPT-VLA~\citep{tan2025interactive} also studies online RL for VLAs. Offline RL methods such as IQL~\citep{kostrikov2022iql} improve a policy using a fixed dataset. CSO instead alternates online collection and skill-level updates after supervised alignment. Our IQL comparison matches the initial checkpoint, skill interface, and collection budget, and compares fixed-data with adaptive-online refinement strategies (Section~\ref{sec:controlled}).

\section{Preliminaries}
\label{sec:preliminary}

We develop a post-training method that refines imitation-learned policies through online interaction. We first define this base policy architecture and then formulate the online policy optimization problem.

\subsection{Skill-Based Policies via Action Discretization}
\label{sec:vla_base}

We consider a base policy trained on an expert dataset $\mathcal{D}$, containing trajectories of observation-action pairs $\{(O_t, a_t)\}_{t=1}^T$ and a language instruction $L$. An observation $O_t$ includes multi-view RGB images and proprioceptive states, while an action $a_t$ is a continuous vector. Following prior skill-based policies such as QueST~\citep{mete2024quest}, the policy is trained in two stages.

\paragraph{1. Skill Codebook Learning.} A quantized autoencoder learns a discrete vocabulary of skills. An encoder $\phi_\theta$ maps a continuous action chunk $a_{t:t+k-1}$ to a latent representation, which is then discretized by a quantizer (e.g., FSQ~\citep{fsq}) into a sequence of skill tokens $Z = (z^1, \ldots, z^n)$. A decoder $\psi_\theta$ reconstructs the action chunk $\hat{a}_{t:t+k-1}$ from $Z$. The autoencoder is trained by minimizing the reconstruction loss:
\begin{equation}
    \mathcal{L}_{\text{recon}}(\theta) = \|\psi_\theta(\text{Quantize}(\phi_\theta(a_{t:t+k-1}))) - a_{t:t+k-1}\|_1.
    \label{eq:recon}
\end{equation}

\paragraph{2. Skill Prior Learning.} A multimodal autoregressive model, the skill prior $\pi_\varphi$, is trained via imitation learning (IL) to predict the expert's skill token sequence $Z_{\text{expert}}$. In our framework, this module can be instantiated either as a lightweight Transformer or as a larger VLM backbone such as Qwen3-VL-2B. It is conditioned on the observation history $\mathcal{O}_{t-h:t}$ and the task embedding $e$. The training objective is to minimize the negative log-likelihood:
\begin{equation}
    \mathcal{L}_\mathrm{IL}(\varphi) = -\log \pi_\varphi(Z_{\text{expert}} | \mathcal{O}_{t-h:t}, e).
    \label{eq:il}
\end{equation}
During inference, the policy $\pi_\varphi$ generates a skill $\hat{Z}$, which the frozen decoder $\psi_\theta$ translates into executable continuous actions.

\subsection{Online Refinement via Policy Optimization}
\label{sec:rl_preliminary}

To improve the IL policy's task success rate, we frame the online refinement as a Partially Observable Markov Decision Process (POMDP). The policy $\pi_\varphi$ maps a state $s_t$, comprising the observation history $\mathcal{O}_{t-h:t}$ and task instruction $e$, to a discrete action, which is a full skill (token sequence) $Z_t$. The environment returns a sparse, binary reward $R \in \{0, 1\}$ upon task completion. The objective is to maximize the expected return:
\begin{equation}
    \max_{\varphi} \mathbb{E}_{\tau \sim \pi_\varphi} [R(\tau)],
    \label{eq:rl_objective}
\end{equation}
where $\tau$ is a trajectory generated by $\pi_\varphi$. We build upon PPO variants for sequence modeling, including Group Sequence Policy Optimization (GSPO)~\citep{zheng2025groupsequencepolicyoptimization}. In our sparse-reward setting, we use a mean-centered critic-free advantage relative to a group $\mathcal{G}$ of trajectories from the same initial state:
\begin{equation}
    A(\tau^{(i)}) = R(\tau^{(i)}) - \bar{R}(\mathcal{G}).
    \label{eq:grpo_advantage}
\end{equation}
The core idea of GSPO is to compute the importance sampling ratio over all skill tokens in an episode, denoted by $\hat{Z}_\tau$. The PPO-clipped objective is:
\begin{equation}
    L_{\text{GSPO}}(\varphi) = \mathbb{E}_{\mathcal{G} \sim \pi_{\varphi_{\text{old}}}} \left[ \sum_{\tau \in \mathcal{G}} \min \left( \rho_{\mathrm{GSPO}}(\hat{Z}_\tau) A(\tau), \text{clip}(\rho_{\mathrm{GSPO}}(\hat{Z}_\tau), 1-\epsilon_{-}, 1+\epsilon_{+}) A(\tau) \right) \right],
    \label{eq:gspo}
\end{equation}
Here $\hat Z_\tau$ denotes all generated tokens in an episode. In the original GSPO formulation, the sequence ratio is length-normalized:
\begin{equation}
 \rho_{\mathrm{GSPO}}(\hat Z_\tau)
 =\exp\!\left[\frac{1}{N_\tau}\sum_{t,j}
 \log\frac{\pi_\varphi(z_t^j\mid s_t,z_t^{<j})}
 {\pi_{\varphi_{\mathrm{old}}}(z_t^j\mid s_t,z_t^{<j})}\right],
 \label{eq:gspo_ratio}
\end{equation}
where $N_\tau$ is the number of generated tokens. Applied at episode level, a single sequence ratio shares one clipping decision across the episode. SPO instead clips each environment-level skill independently. The product of skill likelihoods defines the policy likelihood conditioned on the observed states.

\section{Method}
\label{sec:method}
\begin{figure}[t]
    \centering
    \includegraphics[width=\textwidth]{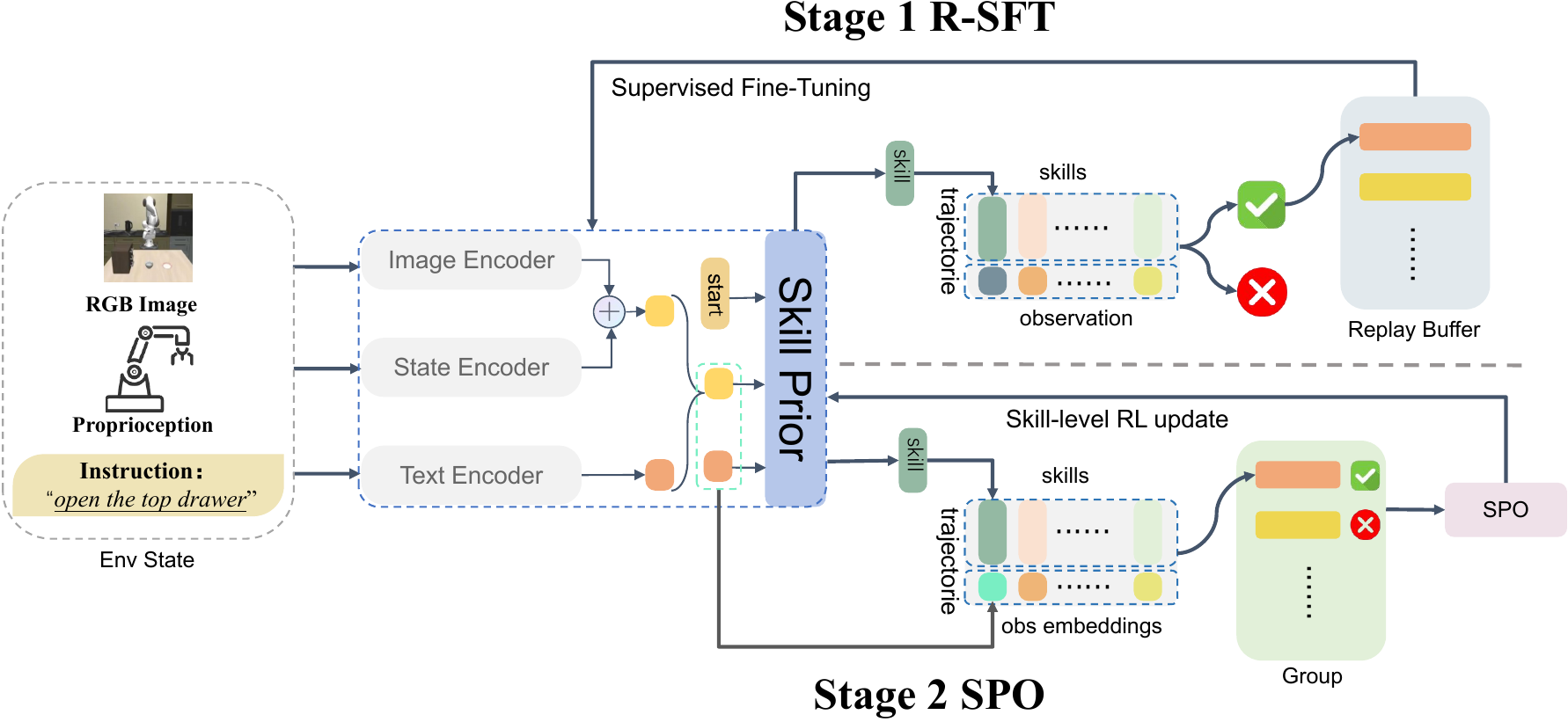}
    \caption{
        The architecture and data flow of our proposed \textbf{\modelacr} framework. 
        \textbf{Stage 1 (R-SFT)}: The initial policy is executed to collect trajectories, and successful trajectories are stored in a replay buffer. The observation-to-skill policy is then fine-tuned on this curated dataset using a supervised objective; the action tokenizer and decoder remain fixed.
        \textbf{Stage 2 (SPO)}: With the encoders now frozen, the policy collects groups of trajectories. The \sposhort algorithm then updates the skill prior with a skill-level policy gradient based on terminal group-relative advantages.
    }
    \label{fig:cso_framework}
\end{figure}

To overcome the challenges of refining imitation-learned, skill-based policies, we propose \textbf{\modelname} (\textbf{\modelacr}), a two-stage post-training framework. Our approach is designed to systematically address the core issues of encoder instability and the granularity dilemma in policy optimization. First, \textbf{\rsft} (\textbf{\rsftshort}) stabilizes the model's representation by aligning it with a distribution of successful online behaviors. Second, with a robust foundation in place, \textbf{\spo} (\textbf{\sposhort}) resolves the granularity dilemma by performing policy optimization on the natural decision-making unit, i.e., the skill.

\subsection{Stage 1: Stabilizing Representations with \rsft (\rsftshort)}

\begin{algorithm}[t]
\caption{\textbf{\modelacr}: A Two-Stage Framework for Refining Skill-Based Policies}
\label{alg:cso_balanced}
\begin{algorithmic}[1]
    \Statex \textbf{Input:} IL-pretrained policy $\pi_\varphi$, Task distribution $\mathcal{T}$, Hyperparameters
    
    \Statex
    \Statex \textbf{▷ Stage 1: \rsft\ (\rsftshort)}
    \State Initialize success buffer $\mathcal{D}_{\text{succ}} \leftarrow \emptyset$
    \While{$|\mathcal{D}_{\text{succ}}| < N_{\text{SFT}}$}
        \State Sample an initial state $s_0 \sim \mathcal{T}$
        \State Execute policy $\pi_\varphi$ $N_{\text{strikes}}$ times from $s_0$ to get trajectories $\{\tau_i\}$
        \State Collect successful trajectories $\mathcal{D}_{\text{succ\_batch}} \leftarrow \{ \tau_i \mid R(\tau_i) = 1 \}$
        \If{$\mathcal{D}_{\text{succ\_batch}}$ is not empty}
            \State Add $\mathcal{D}_{\text{succ\_batch}}$ to $\mathcal{D}_{\text{succ}}$
        \EndIf
    \EndWhile
    \State Fine-tune $\pi_\varphi$ for $E_{\text{SFT}}$ epochs on $\mathcal{D}_{\text{succ}}$ using IL objective (Eq.~\ref{eq:il})
    \State Freeze the observation encoder of $\pi_\varphi$
    
    \Statex
    \Statex \textbf{▷ Stage 2: \spo\ (\sposhort)}
    \For{iteration = 1 to $M$}
        \State $\pi_{\varphi_{\text{old}}} \leftarrow \pi_\varphi$ and initialize $\mathcal{D}_{\text{rollout}} \leftarrow \emptyset$
        \While{$|\mathcal{D}_{\text{rollout}}| < B$}
            \State Sample a group of $K$ trajectories $\{\tau_k\}$ from a single $s_0 \sim \mathcal{T}$ using $\pi_{\varphi_{\text{old}}}$
            \State Compute advantages $\{A(\tau_k)\}$ for the group (Eq.~\ref{eq:skill_advantage})
            \If{all $A(\tau_k)$ are zero} \textbf{continue} \Comment{Dynamic rejection}
            \EndIf
            \State Add trajectories and their advantages to $\mathcal{D}_{\text{rollout}}$
        \EndWhile
        \State Update the \textbf{skill prior} of $\pi_\varphi$ for $N_{\text{SPO}}$ epochs on $\mathcal{D}_{\text{rollout}}$ using \sposhort loss (Eq.~\ref{eq:spo_objective})
    \EndFor
    \State \textbf{return} Refined policy $\pi_\varphi$
\end{algorithmic}
\end{algorithm}

The initial policy, parameterized by $\varphi_{\text{IL}}$, suffers from quantization errors and distribution shift. With sparse binary rewards, groups that are uniformly successful or unsuccessful have zero relative advantage. Other groups provide variable learning signals; jointly adapting perception and the skill prior can destabilize online refinement. The objective of \rsftshort is to circumvent this problem by first refining the policy parameters from $\varphi_{\text{IL}}$ to a more robust set $\varphi_{\text{\rsftshort}}$ using a stable, supervised signal.

\paragraph{Curated Data Collection for Robust Alignment.}

To build a diverse success dataset $\mathcal{D}_{\text{succ}}$ and mitigate overfitting, we employ a curated sampling protocol. At the action level, we execute a stochastic policy (via high-temperature sampling) to encourage exploration. At the state level, we apply an "N-strikes" rule: an initial state is discarded if all $N$ attempts from it fail, and a new one is sampled. This strategy promotes a broad search for successful trajectories across initial states. The stored labels are the skills actually executed in complete successful trajectories. The action tokenizer and decoder remain fixed throughout R-SFT and SPO.

\paragraph{Supervised Alignment and Encoder Stabilization.}
With the curated dataset $\mathcal{D}_{\text{succ}}$ of successful state-skill pairs $\{(s_t, Z_t)\}$, we refine the policy parameters by minimizing the imitation learning objective on this new data distribution:
\begin{equation}
    \varphi_{\text{\rsftshort}} = \arg \min_{\varphi} \mathbb{E}_{(s_t, Z_t) \sim \mathcal{D}_{\text{succ}}} [-\log \pi_\varphi(Z_t | s_t)].
    \label{eq:rsft_objective}
\end{equation}
Upon completion, we freeze the parameters of the observation encoder. Let the policy $\pi_\varphi$ be a composition of an encoder $f_{\text{enc}}$ and a skill prior $h_{\text{prior}}$. Freezing $f_{\text{enc}}$ is critical for providing a stationary observation representation to the skill prior during subsequent RL. The output is a policy $\pi_{\varphi_{\text{\rsftshort}}}$ with a stable encoder, providing a robust foundation for optimization.

\subsection{Stage 2: Resolving the Granularity Dilemma with \spo (\sposhort)}

With a stable representation, we introduce \textbf{\spo} (\textbf{\sposhort}) to effectively optimize the unfrozen \textbf{skill prior}. \sposhort is designed to resolve the granularity dilemma by choosing the skill as the optimization unit between internal tokens and complete trajectories.

\paragraph{Dynamic rollout sampling.}
For online data collection, groups of trajectories from trivially solved or consistently failed initial states produce all-zero advantages. To avoid zero-advantage groups, we employ a dynamic rejection strategy (Algorithm~\ref{alg:cso_balanced}): any group of trajectories where all rollouts receive the same reward is discarded, and a new initial state is sampled. This focuses optimization on tasks at the frontier of the policy's capability.

\paragraph{The Granularity Spectrum.}
Let us consider a trajectory $\tau$ where the policy generates a sequence of skills $(Z_0, Z_1, \ldots, Z_H)$. The policy likelihood on its observed states factors as $\prod_{t=0}^H \pi_\varphi(Z_t | s_t)$. This conditional policy likelihood is evaluated on states sampled through environment rollouts.
\begin{itemize}
    \item \textbf{Token-level optimization} would perform updates based on ratios $\rho(z_t^j)$, clipping internal token decisions separately.
    \item \textbf{Trajectory-level optimization} uses a single ratio $\rho(\tau)$, sharing a clipping decision across an episode whose number of decisions varies between short successes and long time-outs.
\end{itemize}

\paragraph{Skill-Level Policy Optimization.}
\sposhort navigates this spectrum by aligning the optimization unit with the decision unit: the skill. We treat each skill generation at timestep $t$ as a distinct action. The probability of a skill $Z_t = (z_t^1, \dots, z_t^n)$ is:
\begin{equation}
    \pi_\varphi(Z_t | s_t) = \prod_{j=1}^{n} \pi_\varphi(z_t^j | s_t, z_t^{<j}).
    \label{eq:skill_prob}
\end{equation}
Given the sparse reward, the advantage for any skill $Z_t$ within a trajectory $\tau^{(i)}$ is defined relative to the trajectory group's average reward:
\begin{equation}
    A(Z_t) = A(\tau^{(i)}) = R(\tau^{(i)}) - \bar{R}(\mathcal{G}), \quad \forall Z_t \in \tau^{(i)},
    \label{eq:skill_advantage}
\end{equation}
where $\bar{R}(\mathcal{G})$ is the average reward of the trajectory group $\mathcal{G}$.

Combining these elements, the \sposhort objective function updates the unfrozen skill prior by maximizing the following PPO-style clipped surrogate objective:
\begin{equation}
    L_{\text{\sposhort}}(\varphi) = \mathbb{E}_{\tau \sim \pi_{\varphi_{\text{old}}}} \left[ \sum_{t=0}^H \min \left( \rho(Z_t) A(\tau), \text{clip}(\rho(Z_t), 1-\epsilon_{-}, 1+\epsilon_{+}) A(\tau) \right) \right],
    \label{eq:spo_objective}
\end{equation}
where $\rho(Z_t) = \pi_\varphi(Z_t | s_t)/\pi_{\varphi_{\text{old}}}(Z_t | s_t)$. We allow asymmetric clipping bounds; the default SPO setting uses $\epsilon_{-}=0.07$ and $\epsilon_{+}=0.15$.

By computing and clipping an importance ratio for each skill independently, SPO assigns a separate importance weight to each decision while sharing the trajectory's terminal advantage. This skill-aligned update enables stable and data-efficient policy improvement, completing the CSO framework.

\section{Experiments}
\label{sec:experiments}

\begin{table}[tbp]
    \centering
    \small 
    \setlength{\tabcolsep}{4pt} 
    \sisetup{
        detect-weight=true, 
        detect-family=true,
        table-align-uncertainty=true, 
        uncertainty-mode=separate      
    }
    \caption{
        Success rates (\%) on LIBERO. Overall is the unweighted mean of the four displayed suites.
        Baselines marked with $^\dag$ are cited from their original papers.
        Reported uncertainties are standard deviations over three random seeds. Methods are grouped by architectural paradigm.
        The \best{best} and \secondbest{second-best} results are highlighted.
    }
    \label{tab:libero_overall_performance_1}
    \resizebox{\textwidth}{!}{
    \begin{tabular}{l c c c c c}
    \toprule
    {Method} & {LIBERO-Object} & {LIBERO-Spatial} & {LIBERO-Goal} & {LIBERO-Long} & {Overall} \\
    \midrule
    Octo$^\dag$~\citep{octo}      & $85.7 \pm 0.9$ & $78.9 \pm 1.0$ & $84.6 \pm 0.9$ & $51.1 \pm 1.3$ & $75.1 \pm 0.6$ \\
    OpenVLA$^\dag$~\citep{kim2024openvla}    & $88.4 \pm 0.8$ & $84.7 \pm 0.9$ & $79.2 \pm 1.0$ & $53.7 \pm 1.3$ & $76.5 \pm 0.6$ \\
    Dita$^\dag$~\citep{hou2025dita}             & $96.3 \pm 0.7$ & $84.2 \pm 1.0$ & $85.4 \pm 1.5$ & $63.8 \pm 0.8$ & $82.4 \pm 0.8$ \\
    $\pi_0$~\citep{black2024pi0}          & \multicolumn{1}{c}{\secondbest{$98.8 \pm 0.5$}} & $96.8 \pm 0.8$ & $95.8 \pm 1.0$ & $85.2 \pm 1.1$ & $94.2 \pm 0.6$ \\
    $\pi_0$ + FAST$^\dag$~\citep{fast}   & $96.8 \pm 0.7$ & $96.4 \pm 0.9$ & $88.6 \pm 0.8$ & $60.2 \pm 0.2$ & $85.5$ \\
    \midrule
    ResNet-T~\citep{mete2024quest}         & $78.9 \pm 1.4$ & $75.7 \pm 1.9$ & $52.7 \pm 2.4$ & $45.0 \pm 1.1$ & $63.1$ \\
    Diffusion Policy~\citep{chi2023diffusion} & $62.6 \pm 2.8$ & $69.5 \pm 1.8$ & $54.6 \pm 0.5$ & $51.2 \pm 3.0$ & $59.5$ \\
    ACT~\citep{act}              & $78.8 \pm 1.2$ & $82.0 \pm 0.5$ & $66.1 \pm 1.6$ & $44.0 \pm 0.5$ & $67.7$ \\
    MDT~\citep{reuss2024multimodal}              & $87.5 \pm 1.4$ & $78.5 \pm 1.5$ & $73.5 \pm 1.8$ & $64.8 \pm 1.1$ & $76.1 \pm 1.4$ \\
    VQ-BeT~\citep{vq-bet}           & $90.3 \pm 1.5$ & $88.7 \pm 2.0$ & $61.3 \pm 1.0$ & $59.7 \pm 0.2$ & $75.0$ \\
    QueST~\citep{mete2024quest}            & $90.0 \pm 1.1$ & $84.5 \pm 0.2$ & $76.7 \pm 0.9$ & $69.1 \pm 1.0$ & $80.1$ \\
    STAR$^\dag$~\citep{li2025star}             & $98.3 \pm 0.2$ & $95.5 \pm 0.6$ & $95.0 \pm 0.7$ & $88.5 \pm 0.3$ & $94.3$ \\
    \midrule
    GRAPE$^\dag$~\citep{zhang2024grape}            & $92.1 \pm 0.6$ & $88.5 \pm 0.8$ & $83.1 \pm 1.3$ & $57.2 \pm 1.4$ & $80.2 \pm 0.6$ \\
    QueST + RIPT~\citep{tan2025interactive}     & $98.4 \pm 0.3$ & $95.6 \pm 0.4$ & $92.7 \pm 0.5$ & $87.5 \pm 0.6$ & $93.6 \pm 0.2$ \\
    \modelacr{} (QueST-style)             & $98.6 \pm 0.2$ & \secondbest{$98.0 \pm 0.6$} & \secondbest{$97.4 \pm 0.2$} & \secondbest{$96.0 \pm 0.5$} & \secondbest{$97.5 \pm 0.4$} \\
    \modelacr{} (Qwen3-VL-2B)             & \multicolumn{1}{c}{\best{$99.1 \pm 0.1$}} & \multicolumn{1}{c}{\best{$98.3 \pm 0.4$}} & \multicolumn{1}{c}{\best{$99.0 \pm 0.2$}} & \multicolumn{1}{c}{\best{$97.8 \pm 0.7$}} & \multicolumn{1}{c}{\best{$98.6 \pm 0.4$}} \\
    \bottomrule
    \end{tabular}
    }

\end{table}

\subsection{Experimental Setup and Baselines}
We evaluate \modelacr on LIBERO and RoboTwin2.0~\citep{chen2025robotwin2} as our main simulation benchmarks, and further validate the method on real-world robot tasks. Performance is measured by Success Rate (SR), averaged over 50 evaluation episodes per task across three random seeds unless otherwise noted. Simulation results use both the lightweight and Qwen3-VL-2B-based implementations; all physical results use the lightweight implementation. Except for the backbone choice and LoRA adaptation in the Qwen3-VL-2B variant, all other settings are kept the same. Additional MetaWorld ML45 results are deferred to Appendix~\ref{app:multitask}.

We compare against three families of baselines: large-scale VLA models, smaller in-task imitation learning models, and RL-finetuned methods. This setup measures absolute performance and improvement over strong imitation and online-refinement baselines built on comparable backbones. Environment details, baseline implementations, and full training details are deferred to Appendix~\ref{app:environments}, Appendix~\ref{app:baselines}, and Appendix~\ref{app:implementation}.

\subsection{Overall Performance}

As shown in Table~\ref{tab:libero_overall_performance_1}, the QueST-style instantiation of \modelacr achieves 97.5\% overall success on LIBERO, a 17.4-point improvement over its QueST initial policy, with gains in all four suites. On LIBERO-Long, success increases from 69.1\% to 96.0\%, a gain of 26.9 absolute percentage points.

On LIBERO-Long, \modelacr substantially outperforms prior methods, indicating improved robustness to compounding errors in long-horizon tasks. We also observe strong multi-task results on LIBERO-90 (97.7\%) for the QueST-style instantiation. Additional MetaWorld ML45 results are reported in Appendix~\ref{app:multitask}.

\subsection{RoboTwin2.0 Evaluation}
\label{sec:robotwin}

To complement the benchmark results above, we evaluate \modelacr on RoboTwin2.0~\citep{chen2025robotwin2}, a simulated embodiment benchmark that emphasizes visually grounded manipulation under both standard and randomized initializations. We report the main results for the QueST baseline, its RIPT-refined counterpart, and our two implementations in a single table.

\begin{table}[htbp]
    \centering
    \caption{RoboTwin2.0 results. We report success rates (\%) under clean and randomized evaluation settings, together with their average.}
    \label{tab:robotwin_results}
    \small
    \begin{tabular}{l c c c}
        \toprule
        \textbf{Method} & \textbf{Clean (\%)} & \textbf{Randomized (\%)} & \textbf{Average (\%)} \\
        \midrule
        QueST~\citep{mete2024quest} & $67.0 \pm 0.7$ & $65.3 \pm 0.7$ & $66.2 \pm 0.7$ \\
        QueST + RIPT~\citep{tan2025interactive} & $76.6 \pm 0.4$ & $75.3 \pm 0.9$ & $76.0 \pm 0.5$ \\
        \modelacr{} (QueST-style) & \secondbest{$79.6 \pm 1.3$} & \secondbest{$75.7 \pm 0.9$} & \secondbest{$77.7 \pm 0.8$} \\
        \modelacr{} (Qwen3-VL-2B) & \best{$81.2 \pm 1.2$} & \best{$79.6 \pm 0.6$} & \best{$80.4 \pm 0.7$} \\
        \bottomrule
    \end{tabular}
\end{table}

The results in Table~\ref{tab:robotwin_results} show that both CSO implementations outperform the QueST baseline and its RIPT-refined variant, with the Qwen3-VL-2B extension yielding the strongest simulation performance. For the lightweight version, the gains over QueST are 12.6 points in the clean setting and 10.4 points under randomized initialization.

\subsection{Ablation Study on CSO Framework Components}

We conduct ablations on LIBERO-Long and LIBERO-90 to isolate the contribution of each component (Table~\ref{tab:ablation_components}). Both R-SFT and SPO provide substantial gains over the IL baseline, but their combination performs best. This result supports our central design choice of first stabilizing representations and then refining the policy with skill-level optimization. The encoder-freezing ablation further shows the importance of a stable representation during online refinement.

\begin{table}[htbp]
    \centering
    \small
    \setlength{\tabcolsep}{8pt}
    \caption{
        Ablation study on the core components of our \modelacr framework. The synergistic combination of R-SFT and SPO, together with encoder freezing, achieves the strongest performance.
    }
    \label{tab:ablation_components}
    \begin{tabular}{l c c}
        \toprule
        \textbf{Method / Configuration} & \textbf{LIBERO-Long (\%)} & \textbf{LIBERO-90 (\%)} \\
        \midrule
        (1) \quad IL-pretrained policy (Baseline) & 69.1 & 88.6 \\
        \midrule
        (2) \quad \modelacr (Ours, Full Model) & \best{96.0} & \best{97.7} \\
        (3) \quad w/o \sposhort (R-SFT only) & 85.3 & 93.1 \\
        (4) \quad w/o \rsftshort (SPO only) & \secondbest{91.7} & \secondbest{95.3} \\
        (5) \quad w/o Encoder Freeze & 89.6 & 94.4 \\
        \bottomrule
    \end{tabular}
\end{table}

\subsection{Analysis of Granularity in Policy Optimization}
\label{sec:granularity_analysis}

We next analyze the central design choice in \sposhort: skill-level optimization granularity. We compare it against trajectory-level updates (GSPO) and token-level updates (GRPO), with all methods initialized from the same \rsftshort checkpoint.

\begin{figure}[tbp]
 \centering
 \includegraphics[width=0.65\linewidth]{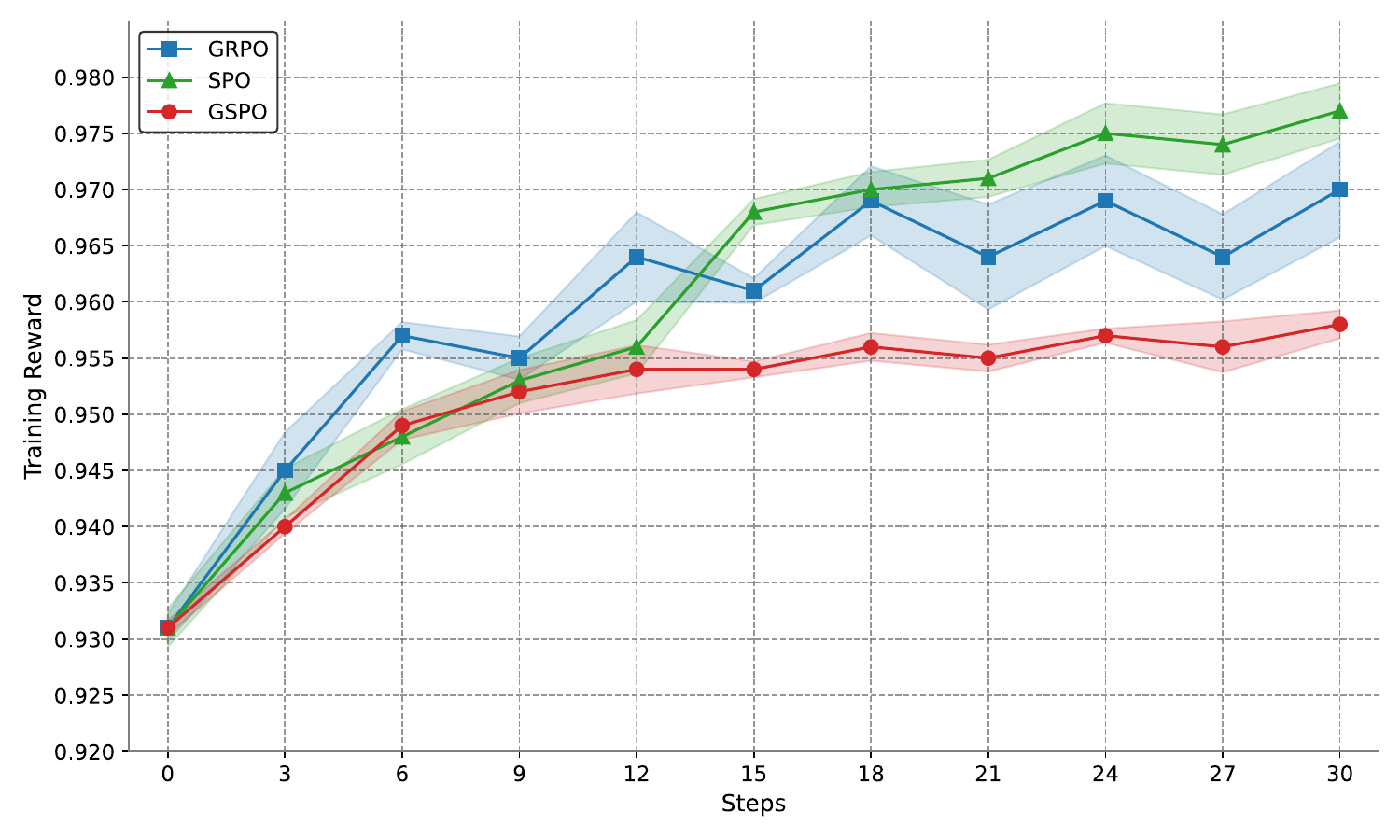}
 \caption{LIBERO-90 learning curves: mean reward with standard-deviation shading over three seeds. Table~\ref{tab:granularity_comparison} separately evaluates LIBERO-Long.}
 \label{fig:granularity_curves}
\end{figure}

Table~\ref{tab:granularity_comparison} compares the best evaluated clipping ranges on LIBERO-Long. SPO reaches 96.0\%, compared with 94.5\% for token-level GRPO, 93.2\% for episode-level GSPO, and 94.3\% for critic-based PPO. Figure~\ref{fig:granularity_curves} shows the separate LIBERO-90 learning comparison. We report the clipping-range sweep in Appendix~\ref{app:clipping_sweep}.

\paragraph{The Mechanics of Granularity: Length Asymmetry.}
Successful trajectories terminate upon task completion, whereas failures often continue until time-out. Episode-level GSPO shares one clipping decision across this variable-length sequence. SPO instead computes and clips a ratio for each fixed-length skill of eight tokens. This gives each environment-level decision its own importance weight and clipping decision, aligning policy optimization with skill execution.

\begin{table}[tbp]
 \centering
 \caption{Stage-2 optimization from the same R-SFT checkpoint on LIBERO-Long. Each method uses its best evaluated clipping range.}
 \label{tab:granularity_comparison}
 \begin{tabular}{l l c}
 \toprule
 Method & Update or credit scheme & SR (\%) \\
 \midrule
 GRPO & Token-level ratio & 94.5 \\
 GSPO & Episode-level ratio & 93.2 \\
 PPO & Learned value critic & 94.3 \\
 SPO & Skill-level ratio & \textbf{96.0} \\
 \bottomrule
 \end{tabular}
\end{table}

\subsection{Controlled Comparisons and Deployment Cost}
\label{sec:controlled}
\paragraph{Representation and data curation.}
Table~\ref{tab:representation_comparison} matches backbones, demonstrations, horizons, budgets, and evaluation. R-SFT and full CSO improve all three representations: stronger reconstruction and execution-validated alignment are complementary. Using the same expert data and collected success set, sequential R-SFT reaches 85.3\%, versus 76.8\% for joint training from scratch and 80.2\% for mixed fine-tuning. Further protocol details appear in Appendix~\ref{app:representations}.

\begin{table}[tbp]
 \centering
 \caption{Matched action representations on LIBERO-Long. Success rates (IL, R-SFT, CSO) are percentages; MSE measures reconstruction.}
 \label{tab:representation_comparison}
 \begin{tabular}{l c c c c}
 \toprule
 Representation & Reconstruction MSE & IL & R-SFT & Full CSO \\
 \midrule
 FSQ/QueST-style & 0.042 & 69.1 & 85.3 & 96.0 \\
 FAST & 0.018 & 78.2 & 89.5 & 97.2 \\
 Discrete BEAST & 0.015 & 84.5 & 91.2 & 98.1 \\
 \bottomrule
 \end{tabular}
\end{table}

\paragraph{Offline RL and total budget.}
Starting from the 85.3\% R-SFT checkpoint, skill-boundary IQL on a fixed buffer of 32,150 successes and 9,050 failures reaches 89.4\%, versus 96.0\% for adaptive online SPO at the same approximately 41,200 attempted-rollout Stage-2 budget. This comparison evaluates fixed-buffer and adaptive-online refinement pipelines. A separate RIPT run matched to CSO's complete approximately 46,580-attempt budget reaches 88.2\%. CSO post-training takes 15.5 wall-clock hours and 62 GPU hours on LIBERO-Long. The physical pipeline takes 21.5 robot hours, approximately 840 attempts, and 127 additional recovery interventions (Appendix~\ref{app:cost_accounting}).

\paragraph{Sensitivity and robustness.}
CSO improves IL policies by more than 25 points at action chunk lengths $k\in\{16,32,64\}$ with eight tokens per skill, although absolute performance varies from 79.8\% to 96.0\%. It also improves policies initialized with 10 or 25 demonstrations per task to 82.4\% or 91.8\%. Under RGB/camera perturbation, proprioceptive noise, and action delay, CSO reaches 88.4\%, 91.2\%, and 84.6\%, versus RIPT's 71.0\%, 76.2\%, and 65.4\% (Appendix~\ref{app:initialization_noise}).

\subsection{Physical Robot Experiments}
\label{sec:real_robot}

\begin{figure}[tbp]
    \centering
    \includegraphics[width=\textwidth]{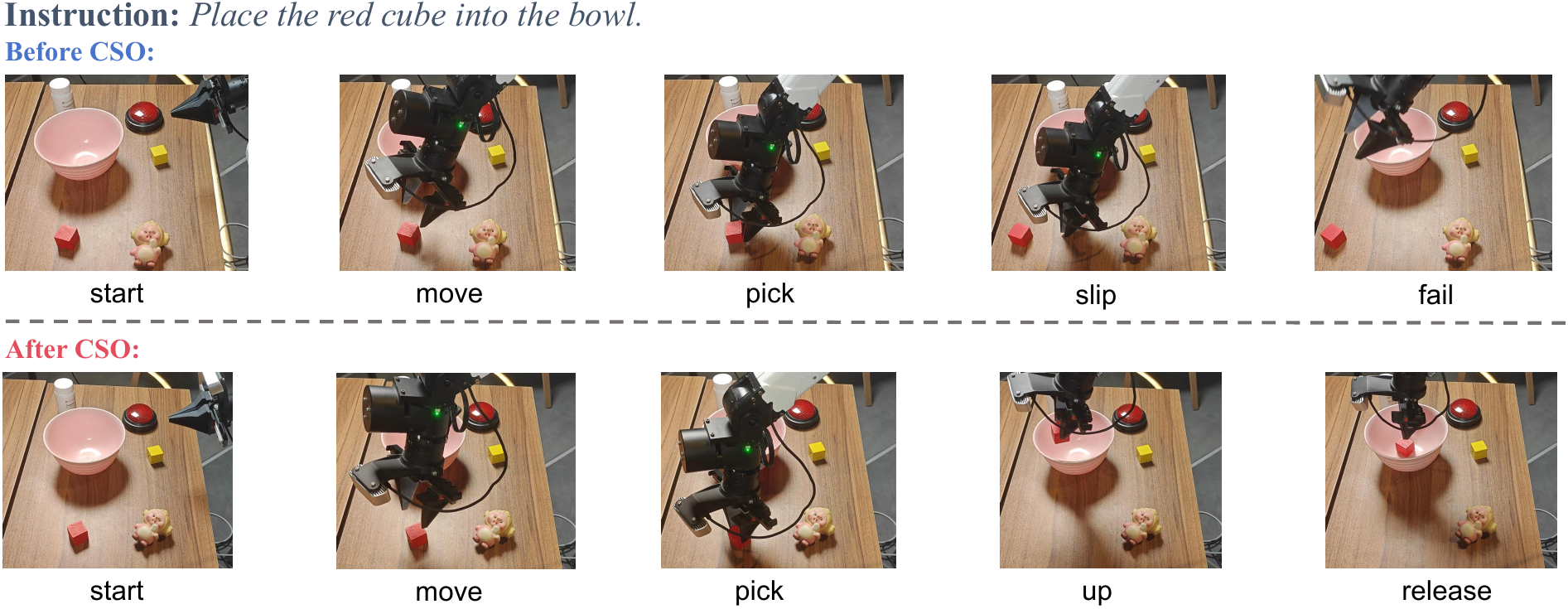} 

    \caption{
        Qualitative results for the \textit{Cube in Bowl} task.
        \textbf{Top (Before \modelacr):} The initial imitation-learned policy fails to grasp the cube due to an incorrect gripper orientation.
        \textbf{Bottom (After \modelacr):} The CSO-refined policy corrects the pose for a successful grasp, demonstrating improved fine-grained control.
    }
    \label{fig:real_world}
\end{figure}

To validate CSO beyond simulation, we evaluate five physical single-arm manipulation tasks against VQ-BeT, QueST, and QueST+RIPT (Appendix~\ref{app:real_world}). As shown in Table~\ref{tab:real_world_all_tasks}, CSO achieves a leading 61.0\% average success rate, a 20-point improvement over QueST. The largest task-wise gain is 30 points on \textit{Press Button}. Figure~\ref{fig:real_world} qualitatively illustrates how CSO corrects the gripper-pose failure in \textit{Cube in Bowl}, demonstrating effective policy refinement on a physical robot.

\begin{table}[tbp]
    \centering
    \small
    \setlength{\tabcolsep}{4pt}
    \caption{Success rates (\%) on five physical tasks, with 20 trials per task.}
    \label{tab:real_world_all_tasks}
    \begin{tabular}{l c c c c c c}
        \toprule
        Method & \shortstack{Cube on\\Cube} & \shortstack{Cube in\\Bowl} & \shortstack{Press\\Button} & \shortstack{Cube on\\Bottle} & \shortstack{Sequential\\Placement} & Average \\
        \midrule
        VQ-BeT & 45 & 35 & 25 & 35 & 5 & 29.0 \\
        QueST & 45 & 50 & 50 & 50 & 10 & 41.0 \\
        QueST + RIPT & 60 & 60 & 70 & \textbf{60} & 25 & 55.0 \\
        CSO & \textbf{65} & \textbf{70} & \textbf{80} & \textbf{60} & \textbf{30} & \textbf{61.0} \\
        \bottomrule
    \end{tabular}
\end{table}

\FloatBarrier
\paragraph{Physical failure and stage-wise analysis.}
Of 39 failed CSO trials, 14 involve configuration/skill selection, 11 tokenizer precision or missing motion, eight contact/grasp/slip, four subgoal order, and two hardware/safety interventions. On \textit{Sequential Placement}, first-stage success is 60\% for CSO versus 50\% for RIPT; both achieve 50\% conditional second-stage success (Appendix~\ref{app:stagewise}). The overall 30\% versus 25\% reflects the higher first-subgoal completion rate.

\section{Conclusion}
We introduced \modelacr, a two-stage framework that refines skill-based robotic policies by decoupling representation stabilization from policy optimization. \modelacr effectively addresses encoder instability and the granularity dilemma. Across LIBERO, RoboTwin2.0, and physical tasks, it consistently outperforms strong baselines, demonstrating that this decomposition is practically effective and well-suited for skill-based control. Additional comparisons with stronger action representations and fixed-buffer IQL further validate the framework.

\section{Limitations and Future Work}
\label{sec:limitations}
\modelacr requires initial success support for each task to bootstrap R-SFT. It remains effective with as few as 50 successful online trajectories per task (Appendix~\ref{app:robustness}); offline RL warm-starting or human correction could extend it to harder exploration settings. The frozen tokenizer's precision and motion coverage, along with the choice of temporal parameters, bound refinement. Physical validation covers five single-arm tasks; broader manipulation settings, automated resets, and sim-to-real transfer offer directions for extending deployment and reducing online collection cost (Appendices~\ref{app:physical_failures} and~\ref{app:cost_accounting}).

\Needspace{12\baselineskip}
\renewcommand{\acksection}{\section*{Acknowledgements}}
\begin{ack}
We would like to thank all co-authors for their efforts and the reviewers for their constructive comments. This work is supported by National Natural Science Foundation of China (Grant No. 62406092), National Natural Science Foundation of China (Grant No. U24B20175), Shenzhen Science and Technology Program (Grant No. KJZD20240903100017022), Guangdong Basic and Applied Basic Research Foundation (Grant No. 2025A1515010169), and Shenzhen Science and Technology Program (Grant No. KQTD20240729102207002).

\textbf{Competing interests.} The authors declare no competing financial interests related to this work during the 36 months prior to submission.

\end{ack}

\bibliographystyle{plainnat}
\bibliography{references}

\clearpage

\appendix
\section*{Appendix}
The appendix provides benchmark and implementation details, additional controlled comparisons, physical failure analysis, interaction and compute accounting, and discussion of the method's scope.  

\section{LLM Usage}

During manuscript preparation, Google's Gemini 2.5 Pro and other large language models were used for editorial assistance, including grammar, organization, and clarity. Separately, the Qwen3-VL-2B simulation variant uses a vision-language model as its multimodal skill prior; its architecture and adaptation are described in Appendix~\ref{app:implementation}. The authors are responsible for the research ideas, implementation, reported results, references, and final text.

\section{Benchmark Environments}
\label{app:environments}
We evaluate CSO across multiple simulation benchmarks to test its effectiveness in diverse manipulation scenarios.

\subsection{LIBERO Benchmark}
LIBERO is a comprehensive benchmark for language-conditioned manipulation tasks that evaluates different aspects of robotic manipulation capabilities through five distinct task suites:
\begin{itemize}
    \item \textbf{LIBERO-Spatial}: Contains 10 tasks focusing on identical objects in different spatial layouts, testing the understanding of spatial relationships.
    \item \textbf{LIBERO-Object}: Comprises 10 tasks with consistent layouts but varying objects, evaluating the ability to generalize across different object types.
    \item \textbf{LIBERO-Goal}: Features 10 tasks sharing the same object categories and spatial layouts but with different goals, assessing the capability to learn diverse task-oriented behaviors.
    \item \textbf{LIBERO-Long}: Presents 10 challenging long-horizon tasks involving diverse object categories and layouts, testing temporal reasoning and sequential manipulation skills.
    \item \textbf{LIBERO-90}: Encompasses 90 tasks with extremely diverse object categories, layouts, and task goals, providing a comprehensive evaluation platform.
\end{itemize}
For our experiments, we utilize 50 expert demonstrations per task from the official dataset. Each demonstration includes multi-view RGB images (224x224 resolution), proprioceptive states (7-dimensional joint positions and 1-dimensional gripper state), and corresponding continuous action commands.

\subsection{MetaWorld ML45}
MetaWorld ML45~\citep{yu2020metaworld} is a challenging multi-task learning benchmark designed to evaluate the ability of a single policy to master various manipulation skills simultaneously. The ML45 split comprises 45 training tasks (ML45-train) and 5 unseen test tasks (ML45-test). Our experiments focus on ML45-train, which requires a \textbf{single multi-task policy} to learn to solve all 45 distinct manipulation tasks.

These tasks are performed by a Sawyer robotic arm in simulation and cover a wide range of skill types, such as:
\begin{itemize}
    \item \textbf{Simple Movements}: \texttt{reach-v2}, \texttt{push-v2}
    \item \textbf{Pick and Place}: \texttt{pick-place-v2}
    \item \textbf{Interaction with Articulated Objects}: \texttt{door-open-v2}, \texttt{drawer-close-v2}, \texttt{window-open-v2}
    \item \textbf{Tool Usage}: \texttt{hammer-v2}, \texttt{stick-push-v2}
    \item \textbf{Fine Manipulation}: \texttt{button-press-topdown-v2}, \texttt{peg-unplug-side-v2}
\end{itemize}
Each task features randomized target and object initial positions, demanding strong generalization capabilities from the policy. The core challenge of ML45 lies in the policy's ability to infer the correct behavior from task instructions (e.g., "close the drawer") and solve all 45 tasks within a shared model parameter space, avoiding interference between tasks.

For training, we use scripted expert policies provided in the official MetaWorld codebase to collect 100 expert demonstrations per task.

\subsection{RoboTwin2.0 Benchmark}
\label{app:robotwin}
RoboTwin2.0~\citep{chen2025robotwin2} is a simulator benchmark focused on visually grounded manipulation and embodiment-sensitive interaction tasks. We use it as a complementary testbed to evaluate whether policy improvements transfer beyond standard offline imitation-learning benchmarks. Following the common protocol, we report task success under both a \textbf{clean} evaluation setting and a more challenging \textbf{randomized} setting with perturbed initial configurations.

\section{Real-World Setup}
\label{app:real_world}
To validate the effectiveness of our approach in real-world scenarios, we used the \textbf{Cobot Agilex ALOHA robot}, a dual-arm manipulator platform. In our experiments, we exclusively used its \textbf{single arm} for manipulation, to maintain consistency with the single-arm setup in simulation environments. We designed five distinct manipulation tasks, described below. For each task, we collected 50 demonstrations through human teleoperation.

\begin{itemize}
    \item \textbf{Cube on Cube}: Stack the yellow cube on top of the red cube.
    \item \textbf{Cube in Bowl}: Place the red cube into the bowl.
    \item \textbf{Press Button}: Press the red button on the table.
    \item \textbf{Cube on Bottle}: Balance the red cube on top of a medicine bottle.
    \item \textbf{Sequential Placement}: A long-horizon task requiring the policy to first place the yellow cube into the bowl, and then move the bowl onto the plate.
\end{itemize}

\section{Evaluation Protocol and Detailed Results}
\label{app:evaluation}

\subsection{Simulated Benchmarks}
For the LIBERO and MetaWorld ML45 benchmarks, we use the \textbf{Success Rate (SR)} as the evaluation metric. The success rate for each task is calculated as the average over 50 evaluation episodes, and we repeat experiments with three different random seeds to report the mean and standard deviation. For RoboTwin2.0, we report the mean task success rate under both clean and randomized settings, together with their average.

\subsection{Real-World Tasks}
For the real-world tasks, we conducted 20 trials per task and reported the final success rate. A trial was successful when all task stages were completed in the correct sequence.

\subsection{Additional Multitask Results}
\label{app:multitask}
\begin{figure}[tbp]
    \centering
    \begin{subfigure}[b]{0.49\textwidth}
        \centering
        \includegraphics[width=0.75\textwidth]{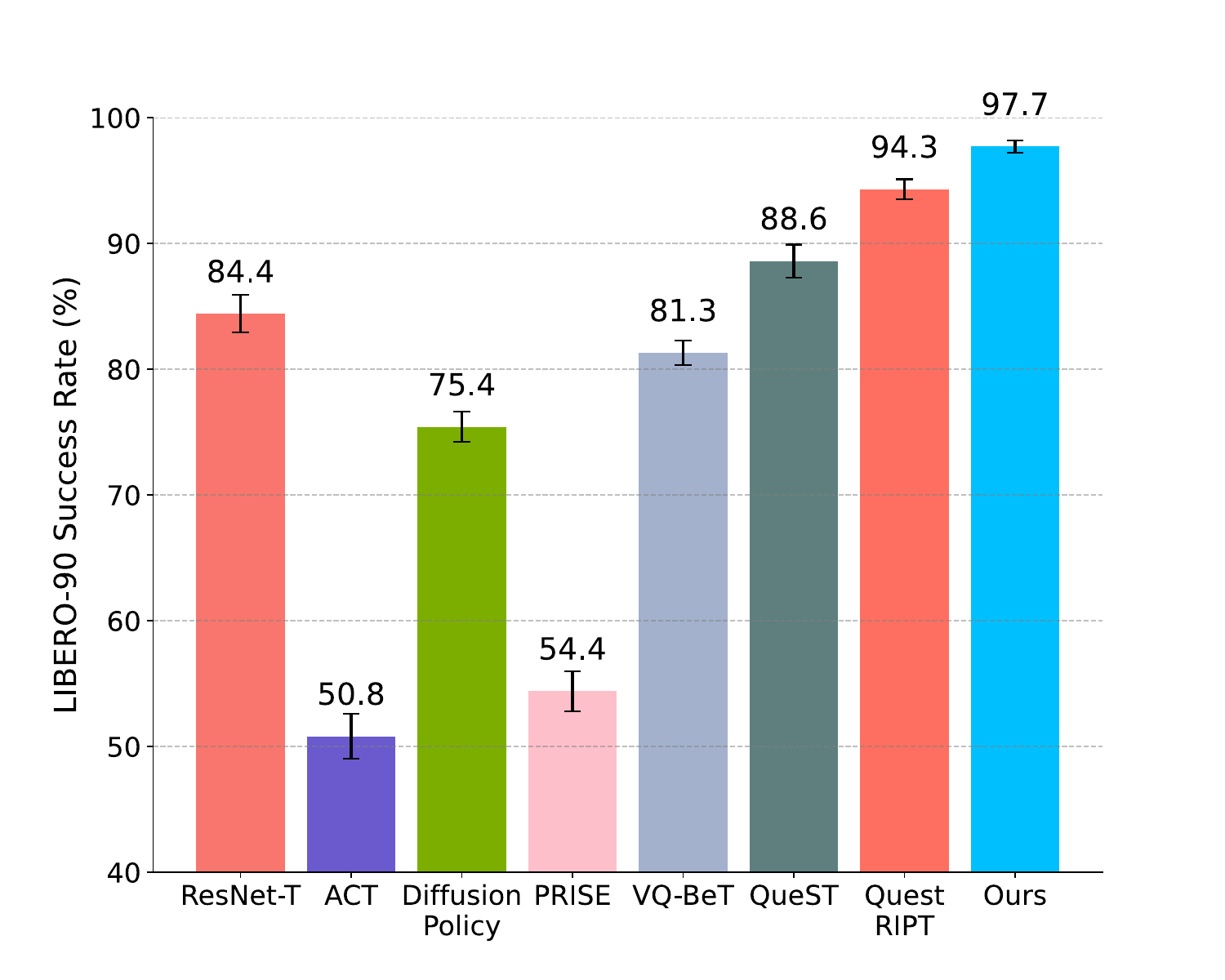}
        \captionsetup{font=scriptsize}
        \caption{Multitask on LIBERO 90}
        \label{fig:libero_90}
    \end{subfigure}
    \hfill
    \begin{subfigure}[b]{0.49\textwidth}
        \centering
        \includegraphics[width=0.75\textwidth]{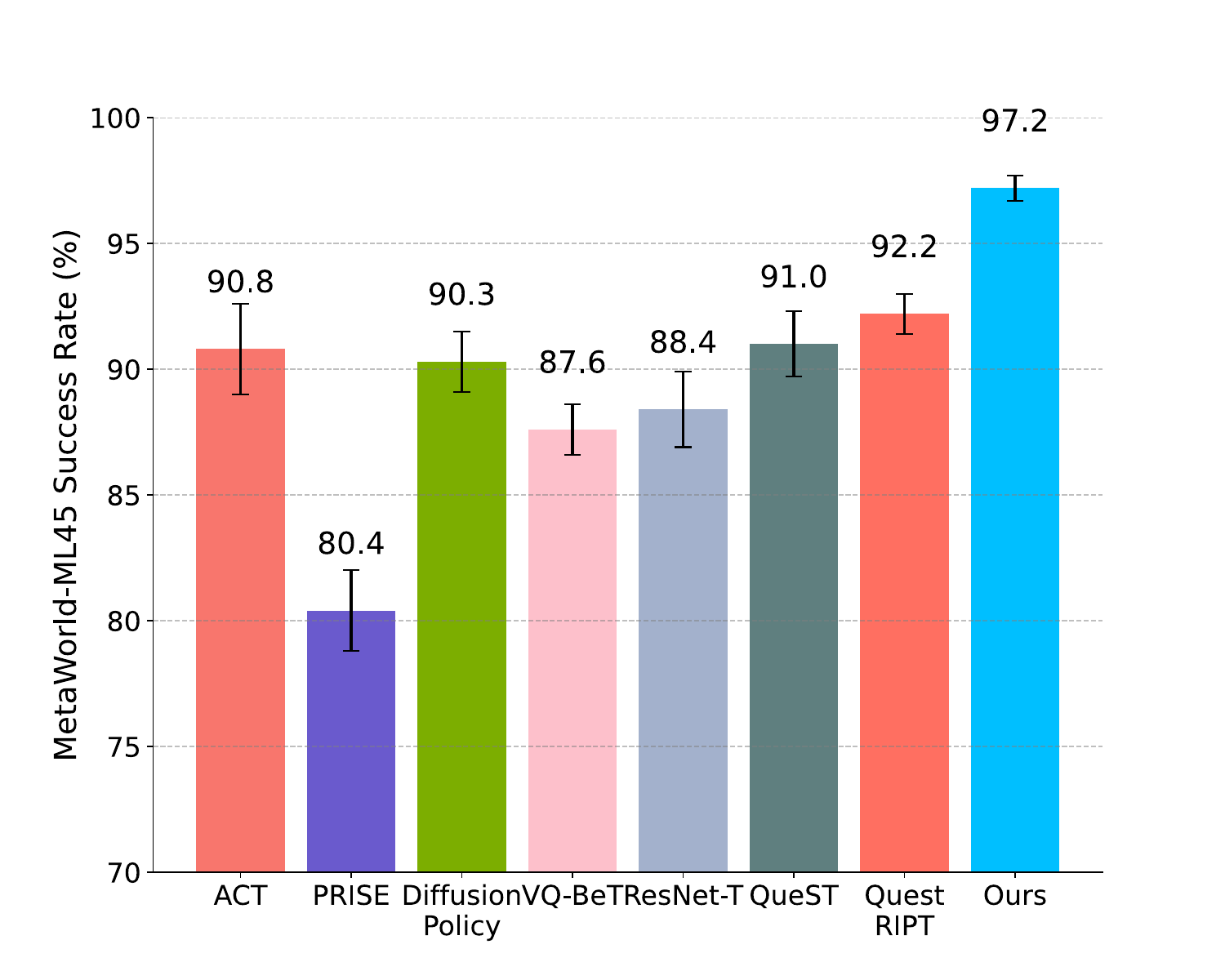}
        \captionsetup{font=scriptsize}
        \caption{Multitask on Metaworld ML45}
        \label{fig:metaworld_ml45}
    \end{subfigure}
    \captionsetup{font=footnotesize}
    \caption{Multitask performance on the LIBERO 90 and Metaworld ML45 benchmarks. Results show the mean and standard error across three random seeds.}
    \label{fig:multitask_performance}
\end{figure}

To complement the aggregate results in Table~\ref{tab:libero_overall_performance_1}, we report the multi-task performance curves on LIBERO-90 and MetaWorld ML45 in Figure~\ref{fig:multitask_performance}. These additional results show that the gains of \modelacr{} extend beyond the four suites in the main performance table and remain consistent in broad multi-task settings.

\subsection{Stage-wise Analysis on the "Sequential Placement" Task}
\label{app:stagewise}
To conduct an in-depth analysis of how different methods perform in long-horizon tasks, we performed a stage-wise success rate analysis for the most challenging real-world task: \textbf{Sequential Placement}. This task requires the robot to first place the yellow cube into the bowl (Stage 1), and then move the bowl onto the plate (Stage 2). The results are presented in Table \ref{tab:real_world_stage_wise}.

\begin{table}[tbp]
    \centering
    \caption{Stage-wise Success Rate (\%) analysis for the real-world "Sequential Placement" task.}
    \resizebox{\textwidth}{!}{%
    \label{tab:real_world_stage_wise}
    \begin{tabular}{l c c c}
        \toprule
        \textbf{Method} & \textbf{Stage 1 Success Rate} & \textbf{Stage 2 Success Rate (Conditional)} & \textbf{Overall Success Rate} \\
        & (Cube into Bowl) & (Bowl onto Plate | Stage 1 Success) & \\
        \midrule
        VQ-BeT (IL Baseline)     & 20\% & 25\% & 5\% \\
        QueST (IL Baseline)      & 40\% & 25\% & 10\% \\
        QueST + RIPT (RL Baseline) & \secondbest{50\%} & \best{50\%} & \secondbest{25\%} \\
        \modelacr (Ours) & \best{60\%} & \best{50\%} & \best{30\%} \\
        \bottomrule
    \end{tabular}
    }
\end{table}

\textbf{Analysis}: Table~\ref{tab:real_world_stage_wise} separates first-subgoal completion from second-stage performance conditioned on first-stage success. VQ-BeT and QueST achieve first-stage success rates of 20\% and 40\%, respectively, followed by 25\% conditional second-stage success. RIPT and CSO both reach 50\% conditional second-stage success, while CSO improves first-stage success from RIPT's 50\% to 60\%. Their overall rates are therefore 25\% and 30\%. In these 20-trial evaluations, the difference corresponds to one additional completed task for CSO. This breakdown locates the observed gain in first-subgoal completion; both methods achieve 50\% conditional second-stage success.

\subsection{Data Collection Cost Analysis}
\label{app:data_cost}
We report the average number of rollouts required to collect the target 200 successful trajectories per task in the R-SFT stage. As shown in Table~\ref{tab:collection_cost}, the number of attempted rollouts exceeds the retained success count and depends on the collection policy and initial-state sampling.

\begin{table}[tbp]
    \centering
    \caption{Average data collection cost to fill the R-SFT buffer (200 successes).}
    \label{tab:collection_cost}
    \begin{tabular}{l c c}
        \toprule
        Benchmark Suite & Initial IL Success Rate & Rollouts Needed (Avg.) \\
        \midrule
        LIBERO-Long & 69.1\% & $\sim$538 \\
        LIBERO-90   & 88.6\% & $\sim$305 \\
        MetaWorld ML45 & 85.0\% & $\sim$341 \\
        \bottomrule
    \end{tabular}
\end{table}

\subsection{Robustness and Design Analysis}
\label{app:robustness}
\textbf{Robustness to Data Scarcity.}
We ablate the number of successful trajectories used in R-SFT on LIBERO-Long. As shown in Table~\ref{tab:robustness_design} (Left), reducing the data from 200 to 50 samples (25\% of default) gives a success rate of \textbf{91.5\%}, which remains above the IL baseline (69.1\%). This indicates that CSO remains effective even in a low-data post-training regime.

\textbf{Two-Stage vs. Iterative Refinement.}
We compare our two-stage design against an iterative loop that alternates between supervised refinement and policy optimization. Results in Table~\ref{tab:robustness_design} (Right) show that iterative refinement yields only a marginal gain (\textbf{+0.3 percentage points}) while requiring \textbf{2.5$\times$} the training time. This supports our decoupled design, suggesting that the main bottleneck lies in policy optimization rather than repeated representation re-alignment.

\begin{table}[tbp]
    \centering
    \caption{Ablations on LIBERO-Long: sample efficiency (left) and design strategy (right).}
    \label{tab:robustness_design}
    \small
    \setlength{\tabcolsep}{4pt}
    \begin{tabular}{cc cc}
        \toprule
        \multicolumn{2}{c}{\textbf{Sample Efficiency (R-SFT)}} & \multicolumn{2}{c}{\textbf{Iterative vs. Two-Stage}} \\
        \midrule
        \textbf{\# Samples} & \textbf{Success Rate (\%)} & \textbf{Strategy} & \textbf{Success Rate (\%)} \\
        \midrule
        50  & 91.5 $\pm$ 0.7 & Two-Stage (Ours) & 96.0 $\pm$ 0.5 \\
        100 & 94.2 $\pm$ 0.5 & Iterative Loop & 96.3 $\pm$ 0.6 \\
        200 & 96.0 $\pm$ 0.5 & \textit{Training Cost} & \textit{2.5$\times$ Time} \\
        \bottomrule
    \end{tabular}
\end{table}

\subsection{Skill Sequence Length Sensitivity}
\label{app:skill_length}
We analyze the effect of skill length (tokens per skill) on LIBERO-Long performance. The token count $n$ controls compression and autoregressive generation difficulty, whereas action chunk length $k$ controls the decision horizon. Table~\ref{tab:skill_len} varies $n$ at the default horizon; eight tokens per skill perform best among the tested values. Table~\ref{tab:action_chunk} separately varies $k$ with $n=8$.

\begin{table}[tbp]
    \centering
    \caption{Effect of Skill Sequence Length on Performance.}
    \label{tab:skill_len}
    \begin{tabular}{c c l}
        \toprule
        Skill Length & Success Rate (\%) & Observation \\
        \midrule
        4 & 93.5 & More aggressive compression. \\
        \textbf{8 (Ours)} & \textbf{96.0} & Best evaluated value. \\
        16 & 91.6 & Longer autoregressive generation. \\
        \bottomrule
    \end{tabular}
\end{table}

\begin{table}[tbp]
 \centering
 \caption{Action chunk length sensitivity on LIBERO-Long at $n=8$ tokens per skill. Gains are absolute percentage points over the corresponding initial policy.}
 \label{tab:action_chunk}
 \begin{tabular}{c c c c}
 \toprule
 Action chunk $k$ & IL SR (\%) & CSO SR (\%) & Gain \\
 \midrule
 16 & 62.4 & 88.5 & +26.1 \\
 32 & 69.1 & 96.0 & +26.9 \\
 64 & 54.2 & 79.8 & +25.6 \\
 \bottomrule
 \end{tabular}
\end{table}
CSO improves all three initial policies by more than 25 points. The best absolute success rate occurs at $k=32$ in this setup. Shorter and longer chunks change the balance between feedback responsiveness and temporal abstraction, so the best horizon may differ across tasks, robot embodiments, and control frequencies. All three settings use fixed-duration skills under the same control and execution protocol.

\subsection{Clipping Hyperparameter Analysis}
\label{app:clipping_sweep}
To ensure a fair comparison in Section~\ref{sec:granularity_analysis}, we performed a grid search for the clipping hyperparameters of baselines. Table~\ref{tab:clip_sweep} shows the results on LIBERO-Long. SPO demonstrates superior robustness across ranges.

\begin{table}[tbp]
    \centering
    \caption{Clipping range sensitivity analysis. \sposhort maintains high performance across ranges, while GSPO is highly sensitive to the clipping bounds.}
    \label{tab:clip_sweep}
    \small
    \begin{tabular}{l c c c}
        \toprule
        Method & Clipping Range & Success Rate & Clipping Fraction \\
        \midrule
        \textbf{Token (GRPO)} & $[0.9, 1.1]$ & 93.8\% & 1.2\% \\
        & $[0.8, 1.2]$ & 94.2\% & 3.1\% \\
        & $[0.8, 1.28]$ (Best) & 94.5\% & 3.5\% \\
        \midrule
        \textbf{Trajectory (GSPO)} & $[0.9997, 1.0004]$ & 92.1\% & 0.8\% \\
        & $[0.997, 1.004]$ (Best) & 93.2\% & 1.5\% \\
        & $[0.97, 1.04]$ & 86.4\% & 12.3\% \\
        \midrule
        \textbf{Skill (SPO, Ours)} & $[0.97, 1.05]$ & 95.5\% & 14.5\% \\
        & $[0.93, 1.15]$ (Best) & \textbf{96.0\%} & 11.2\% \\
        & $[0.8, 1.2]$ & 95.8\% & 5.6\% \\
        \bottomrule
    \end{tabular}
\end{table}

\subsection{Critic-Based vs. Critic-Free Optimization}

\label{app:critic_ablation}
We empirically verified whether a learned value critic (Standard PPO) could improve upon our critic-free SPO approach. Both methods started from the same R-SFT stabilized encoder.

\begin{table}[tbp]
    \centering
    \caption{Comparison of Critic-Free (SPO) vs. Critic-Based (PPO) optimization.}
    \label{tab:critic_vs_free}
    \begin{tabular}{l l c}
        \toprule
        Method & Credit Assignment & LIBERO-Long SR \\
        \midrule
        \textbf{SPO (Ours)} & Critic-free / Shared terminal advantage & \textbf{96.0\%} \\
        Standard PPO & Learned Critic / Fine-grained & 94.3\% \\
        \bottomrule
    \end{tabular}
\end{table}

\section{Implementation Details of CSO}
\label{app:implementation}

Our models are implemented in PyTorch and trained on a server equipped with 4 NVIDIA A100 80GB GPUs.

\subsection{Base Policy Architecture}
Our policy follows a generic skill-based action discretization framework. The core architecture is implementation-agnostic with respect to the skill prior: in this work, we retain a QueST-style lightweight implementation as our small-model version, and we additionally support a Qwen3-VL-2B-based implementation as a large-model extension. In both cases, the overall CSO pipeline remains unchanged. All real-world experiments use the lightweight implementation; simulation experiments use both variants. Other than the backbone choice and the use of LoRA in the Qwen3-VL-2B variant, all training and evaluation settings are shared.

\paragraph{Offline action-tokenizer learning.}
The action tokenizer quantizes continuous action sequences into discrete skill tokens.
\begin{itemize}
    \item \textbf{Architecture:} The encoder consists of 2 Transformer layers with 4 attention heads, and the decoder consists of 4 Transformer layers with 4 attention heads. Both the encoder and decoder have a hidden dimension of 256. We use causal convolutional layers with 3 layers, kernel sizes of $[5,3,3]$, and strides of $[2,2,1]$.
    \item \textbf{Quantization:} We employ Finite Scalar Quantization (FSQ) with quantization levels set to $[8,5,5,5]$.
    \item \textbf{Input:} The input continuous action sequence length $T$ is 32.
\end{itemize}

\paragraph{Offline skill-prior learning.}
\begin{itemize}
    \item \textbf{QueST-style lightweight implementation:} As our small-model instantiation, we use a shallow Convolutional Neural Network (CNN) visual encoder consisting of the first four layers of a ResNet-18 followed by a spatial softmax layer, together with an auto-regressive Transformer that predicts skill tokens from observations. This Transformer has 6 layers, 6 attention heads, an embedding dimension of 384, a vocabulary size of 1000, $n=8$ tokens per skill, and an observation history window of 1. Each skill decodes a $k=32$ continuous action chunk by default; $k$ and $n$ are distinct parameters.
    \item \textbf{Qwen3-VL-2B implementation:} As our large-model instantiation, we replace the lightweight skill prior with Qwen3-VL-2B-Instruct~\citep{bai2025qwen3vl}, an image-text-to-text conditional generation model in the Qwen3-VL family. According to its official configuration, this model uses a 24-layer vision backbone with hidden size 1024, intermediate size 4096, 16 attention heads, patch size 16, temporal patch size 2, and spatial merge size 2, together with a 28-layer text backbone with hidden size 2048, intermediate size 6144, 16 attention heads, 8 key-value heads, and a maximum position length of 262{,}144 tokens. In this variant, Qwen3-VL-2B serves as the multimodal skill prior, while the downstream skill-token prediction interface and CSO optimization procedure remain the same. We train this variant with LoRA of rank 8, inserting LoRA adapters into the attention projection layers of the skill prior. All settings other than the backbone choice and LoRA adaptation are kept the same as in the lightweight implementation. We evaluate this extension in simulation; the physical experiments use the lightweight QueST-style policy.
\end{itemize}

\subsection{CSO Framework Training}
After offline imitation learning, CSO post-training uses two stages. The action tokenizer and decoder are fixed throughout both stages.

\paragraph{Stage 1: Rejection-Sampling SFT (R-SFT).}
\begin{itemize}
    \item \textbf{Data Collection:} For the default simulation setting, we aim to collect 200 successful state-skill trajectories per task (20 per task in the physical setting), which are stored in a replay buffer $\mathcal{D}_{\text{succ}}$. We employ an "N-strikes" sampling rule with $N_{\text{strikes}} = 10$, meaning we attempt up to 10 rollouts from an initial state. If any are successful, all successful trajectories are added to the buffer.
    \item \textbf{Fine-tuning:} Once the buffer is filled, the observation-to-skill policy (including observation encoders, excluding the action tokenizer and decoder) is fine-tuned on $\mathcal{D}_{\text{succ}}$ for 20 epochs. We use the AdamW optimizer with a batch size of 256 and a learning rate that decays from $1\text{e-}4$ to $1\text{e-}5$ using a cosine schedule.
\end{itemize}

\paragraph{Stage 2: Skills Policy Optimization (SPO).}
\begin{itemize}
    \item \textbf{Encoder Freezing:} Upon completion of the R-SFT stage, all parameters of the visual and proprioceptive encoders are \textbf{frozen}.
    \item \textbf{Online Optimization:} We perform a total of 30 iterations. In each iteration, we collect a batch of data using the old policy $\pi_{\varphi_{\text{old}}}$. In the default simulation setting, each batch contains 150 retained groups of eight trajectories from a shared initial state per group. Thirty iterations retain 36,000 trajectories; the approximately 41,200 attempted rollouts also include rejected zero-advantage groups. Physical collection uses the smaller budget in Appendix~\ref{app:cost_accounting}.
    \item \textbf{SPO Update:} The \textbf{unfrozen skill prior} is updated using our SPO objective. In this objective, the importance sampling ratio is clipped to the range $[0.93,1.15]$. We use the AdamW optimizer with a learning rate of $1\text{e-}6$.
\end{itemize}

\subsection{Granularity Ablation Hyperparameters}
In the granularity analysis (Figure~\ref{fig:granularity_curves} in the main paper), we compare our method against other optimization granularities. For a fair comparison, we set the hyperparameters for these baseline algorithms as follows:
\begin{itemize}
    \item \textbf{Token-level (GRPO):} The GRPO comparison uses the importance ratio clipping range $[0.8,1.28]$, which also gives its best LIBERO-Long result in Table~\ref{tab:clip_sweep}.
    \item \textbf{Trajectory-level (GSPO):} The GSPO comparison uses the range $[0.997,1.004]$, which gives its best LIBERO-Long result in Table~\ref{tab:clip_sweep}. The sweep also evaluates the original paper's narrower $[0.9997,1.0004]$ setting~\citep{zheng2025groupsequencepolicyoptimization}.
\end{itemize}

\section{Baseline Implementation Details}
\label{app:baselines}
For fair comparison, all baseline models that we reimplemented utilize the same input modalities as our method. For the small-model setting, our implementation and the reimplemented baselines also share a ResNet-18 visual backbone. We tuned key hyperparameters for the reported comparisons.

\begin{itemize}
    \item \textbf{$\pi_0$ \& $\pi_0$ + FAST}: $\pi_0$~\citep{black2024pi0} uses a vision-language backbone with a flow-matching action expert. FAST~\citep{fast} instead provides compressed discrete action tokens. These large-model baselines are distinct from our controlled tokenizer comparison, which matches the lightweight observation backbone and autoregressive policy interface. Literature-sourced scores are marked with $^\dag$ in Table~\ref{tab:libero_overall_performance_1}.
    \item \textbf{ACT}: The architecture comprises 8 cross-attention and 4 self-attention layers. ACT predicts continuous action chunks using a conditional variational autoencoder. The action chunk length is set to 20.
    \item \textbf{Diffusion Policy}: The backbone employs a U-Net design with channel dimensions $[256, 512, 1024]$. The number of diffusion steps is 100. For the LIBERO benchmark, the prediction and execution horizons are set to $T=32$ and $T_a=16$, respectively.
    \item \textbf{VQ-BeT}: This framework employs a single-layer MLP encoder (dimension 256) and a residual vector quantization module with $K=1024$ codes. The observation window is 5 timesteps, and the action window size is $T=8$.
    \item \textbf{QueST (IL Baseline)}: This serves as the initial policy for our CSO method. Its implementation details are provided in Section \ref{app:implementation}. We report its performance after completing imitation learning on expert data.
    \item \textbf{QueST + RIPT}: This is an important RL fine-tuning baseline. We strictly reimplemented the RIPT algorithm in our codebase and applied it to the exact same QueST initial policy as our CSO method, ensuring the most direct and fair comparison. The primary table uses the Stage-2-matched setting. A separate full-pipeline budget comparison includes the R-SFT collection attempts and is reported in Appendix~\ref{app:cost_accounting}.
\end{itemize}

\section{Collection Protocol and Success Support}
\label{app:rsft_efficiency}
R-SFT stores complete successful trajectories, while every attempted rollout counts toward interaction cost. Under the $N_{\mathrm{strikes}}=10$ protocol, the stochastic policy is executed repeatedly from one initial state before a new state is sampled. The default simulation buffer target is 200 successes per task; LIBERO-Long therefore retains 2,000 successes from approximately 5,380 attempts, an observed retention fraction of about 37.2\%. The 69.1\% initial-policy evaluation score and the 37.2\% stochastic collection retention rate use their respective evaluation and sampling protocols. The success buffer favors configurations solvable by the stochastic collection policy. Successful collection therefore relies on initial policy support for each task.

The physical buffer contains 20 successes per task, collected directly on the robot. Subsequent SPO also uses physical rollouts. The cost measurements cover direct collection on the physical robot.

The representation comparison holds the observation backbone and autoregressive policy interface fixed, while the IQL comparison contrasts fixed-buffer learning with adaptive online refinement from the same R-SFT checkpoint. Stage-level cost accounting covers the measured post-training collection and optimization runs.

\section{Controlled Action-Representation and Data Comparisons}
\label{app:representations}
\subsection{Stronger representations and execution-validated alignment}
We match the observation backbone, policy capacity, expert demonstrations, action horizon, post-training budget, and evaluation protocol on LIBERO-Long. Table~\ref{tab:representation_comparison} compares our FSQ/QueST-style representation with FAST~\citep{fast} and discrete BEAST~\citep{zhou2025beast}. The tokenizer is fixed during R-SFT and SPO in each setting.

R-SFT improves FAST by 11.3 points and discrete BEAST by 6.7 points; full CSO improves their initial policies by 19.0 and 13.6 points. Better reconstruction improves imitation learning, but the decoded expert action remains an approximation. R-SFT trains on state-skill pairs from complete trajectories that succeed under the actual decoded execution pipeline. This execution-level validation complements representation quality while keeping the tokenizer fixed.

FAST and discrete BEAST use discrete action tokens and the same autoregressive policy interface in this controlled comparison. The reconstruction metric is MSE, whereas the FSQ tokenizer training objective in Eq.~\ref{eq:recon} is an $L_1$ loss.

\subsection{Training sequence and data curation}
R-SFT fine-tunes the expert-trained policy on successful online execution trajectories. Encoding and decoding a continuous expert action changes the executed control. Reconstruction errors can affect the task outcome during decoded execution. Static curation may improve the quality or diversity of demonstrations, but execution-level validation requires running the discrete policy and decoder.

Table~\ref{tab:curation_comparison} compares four protocols using the same expert data and collected success set. Joint training starts from scratch using both datasets; mixed fine-tuning starts from the expert-only IL checkpoint and uses both datasets; sequential R-SFT starts from that checkpoint and fine-tunes on the collected successes.

\begin{table}[tbp]
 \centering
 \caption{Data and training-protocol comparison on LIBERO-Long. All alternatives using collected successes have access to the same success set.}
 \label{tab:curation_comparison}
 \begin{tabular}{l c}
 \toprule
 Protocol & SR (\%) \\
 \midrule
 Expert-only IL & 69.1 \\
 Joint training from scratch & 76.8 \\
 Mixed fine-tuning & 80.2 \\
 Sequential R-SFT & 85.3 \\
 \bottomrule
 \end{tabular}
\end{table}
Sequential R-SFT outperforms joint and mixed training by 8.5 and 5.1 points. The protocols use the same collected success set and vary initialization, training sequence, and mixture weighting.

\section{Fixed-Buffer Offline RL Comparison}
\label{app:offline_rl}
We evaluate IQL~\citep{kostrikov2022iql} as a Stage-2 alternative to SPO from the same 85.3\% R-SFT checkpoint. The fixed buffer contains 32,150 successful and 9,050 failed trajectories collected under the same approximately 41,200 attempted-rollout Stage-2 budget. IQL then trains on this fixed buffer. Its critic $Q(s_t,Z_t)$ operates on complete discrete skills at the same decision boundaries as SPO, and the advantage-weighted actor uses the skill's autoregressive sequence likelihood.

\begin{table}[tbp]
 \centering
 \caption{Fixed-buffer and adaptive online refinement on LIBERO-Long, from the same R-SFT initialization and with matched Stage-2 interaction cost.}
 \label{tab:offline_rl}
 \begin{tabular}{l l c}
 \toprule
 Method & Data strategy & SR (\%) \\
 \midrule
 R-SFT only & No Stage-2 optimization & 85.3 \\
 R-SFT + IQL & Fixed offline buffer & 89.4 \\
 R-SFT + SPO & Adaptive online collection & 96.0 \\
 \bottomrule
 \end{tabular}
\end{table}
SPO outperforms IQL by 6.6 points at the matched interaction cost. IQL supports optimization and retuning from the fixed buffer. SPO alternates fresh collection and policy updates. 

\section{Initialization and Perturbation Robustness}
\label{app:initialization_noise}
\subsection{Weaker initial policies}
We vary the number of expert demonstrations used to initialize each LIBERO-Long policy. Stochastic collection produces successful trajectories for every task in all three evaluated settings (Table~\ref{tab:initialization}).

\begin{table}[tbp]
 \centering
 \caption{Refinement of weaker initial policies on LIBERO-Long.}
 \label{tab:initialization}
 \begin{tabular}{c c c}
 \toprule
 Expert demonstrations per task & Initial SR (\%) & Final CSO SR (\%) \\
 \midrule
 10 & 12.0 & 82.4 \\
 25 & 38.5 & 91.8 \\
 50 & 69.1 & 96.0 \\
 \bottomrule
 \end{tabular}
\end{table}
CSO improves the three initial policies from 12.0\%, 38.5\%, and 69.1\% success to 82.4\%, 91.8\%, and 96.0\%, respectively. The separate R-SFT buffer-size ablation in Table~\ref{tab:robustness_design} uses 50, 100, or 200 \emph{successful online trajectories}; the initialization comparison varies the number of \emph{expert demonstrations} used to train the initial policies. With 50 R-SFT successes per task, CSO reaches $91.5\pm0.7$\%.

\subsection{Input perturbations and action delay}
Table~\ref{tab:perturbations} reports success rates under three perturbation conditions. CSO outperforms QueST and its RIPT-refined version in each reported condition. Separately, RoboTwin2.0 randomized evaluation tests initial-configuration shift, where the lightweight CSO policy reaches 75.7\%, compared with 65.3\% for QueST (Table~\ref{tab:robotwin_results}).

\begin{table}[tbp]
 \centering
 \caption{Success rates (\%) under the reported perturbation conditions. Results are point estimates for each perturbation condition.}
 \label{tab:perturbations}
 \begin{tabular}{l c c c}
 \toprule
 Perturbation & QueST & QueST + RIPT & CSO \\
 \midrule
 RGB/camera perturbation & 42.3 & 71.0 & 88.4 \\
 Proprioceptive noise & 51.0 & 76.2 & 91.2 \\
 Action delay & 38.5 & 65.4 & 84.6 \\
 \bottomrule
 \end{tabular}
\end{table}
The three conditions test sensitivity to visual input, proprioception, and action timing.

\section{Physical Failure Analysis and Deployment Scope}
\label{app:physical_failures}
The physical evaluation uses five single-arm tasks with 20 trials each. The shared action tokenizer is trained from 50 teleoperated demonstrations per task. CSO succeeds in 61 of the 100 evaluation trials. Table~\ref{tab:physical_failures} categorizes the 39 unsuccessful trials.

\begin{table}[tbp]
 \centering
 \caption{Failure categories for the 39 unsuccessful physical CSO evaluation trials.}
 \label{tab:physical_failures}
 \begin{tabular}{p{0.78\linewidth} r}
 \toprule
 Failure category & Trials \\
 \midrule
 Unseen configurations or incorrect skill selection & 14 \\
 Insufficient tokenizer precision or missing motion patterns & 11 \\
 Contact, grasp, or slippage errors & 8 \\
 Long-horizon subgoal errors & 4 \\
 Hardware or safety stops & 2 \\
 \midrule
 Total & 39 \\
 \bottomrule
 \end{tabular}
\end{table}
Configuration/skill coverage and tokenizer/motion coverage account for the two largest categories, together 25 of the 39 failures. The frozen tokenizer sets the available skill vocabulary and reconstruction precision, while CSO refines skill selection. The evaluation also records contact errors and hardware interventions. These tests evaluate physical refinement across five single-arm manipulation tasks.

\section{Interaction, Compute, and Physical Cost}
\label{app:cost_accounting}
We distinguish all attempted rollouts from retained trajectories. Stage-2 dynamic rejection discards groups with uniform rewards from SPO's training buffer, but their attempted rollouts still incur cost. Table~\ref{tab:cost_accounting} reports the LIBERO-Long post-training measurements and a separate full-budget RIPT comparison.

\begin{table}[tbp]
 \centering
 \caption{LIBERO-Long post-training cost. GPU hours are aggregated across the four A100 80GB GPUs. RIPT is separately matched to CSO's complete attempted-rollout budget.}
 \label{tab:cost_accounting}
 \begin{tabular}{l r r r}
 \toprule
 Stage or strategy & Attempts & Wall-clock hours & GPU hours \\
 \midrule
 CSO R-SFT & $\sim$5,380 & 3.5 & 14.0 \\
 CSO SPO & $\sim$41,200 & 12.0 & 48.0 \\
 Complete CSO & $\sim$46,580 & 15.5 & 62.0 \\
 RIPT, full budget matched & $\sim$46,580 & 14.5 & 58.0 \\
 \bottomrule
 \end{tabular}
\end{table}
The full-budget RIPT run obtains 88.2\% success, compared with 96.0\% for CSO. It is distinct from the 87.5\% Stage-2-matched RIPT result in Table~\ref{tab:libero_overall_performance_1}. Thus, the main-table comparison and full-pipeline comparison use different total interaction budgets. CSO's 15.5 hours of post-training also represent an additional cost over the initial IL checkpoint. The cost measurements cover the post-training runs listed in the table.

\begin{table}[tbp]
 \centering
 \caption{Physical post-training across all five tasks. Robot time includes routine between-trial scene setup. Additional recoveries count workspace-boundary and safety-stop interventions. GPU hours refer to optimization compute, separately from robot hours.}
 \label{tab:physical_cost}
 \small
 \begin{tabular}{l r p{0.28\linewidth} r r r}
 \toprule
 Stage & Attempts & Retained data & Robot h & GPU h & Recoveries \\
 \midrule
 R-SFT & $\sim$240 & 100 successes (20/task) & 6.5 & 2.0 & 42 \\
 SPO & $\sim$600 & 400 trajectories & 15.0 & 5.0 & 85 \\
 Total & $\sim$840 & 100 successes + 400 trajectories & 21.5 & 7.0 & 127 \\
 \bottomrule
 \end{tabular}
\end{table}
All physical R-SFT and SPO rollouts are collected directly on the robot. Workspace constraints, joint torque limits, emergency stopping, and homing procedures reduce risk and wear during collection. The physical training runs required 127 workspace-boundary and safety-stop recoveries across 21.5 robot hours; the final evaluation recorded two hardware or safety failures among its 39 unsuccessful trials. Future work can examine automated resets, simulation pre-collection, and sim-to-real transfer to reduce collection effort.

\section{Broader Impacts}
\label{app:broader_impacts}
Improving robotic policy refinement may make manipulation more reliable in manufacturing, logistics, and assistive applications. Residual skill-selection, contact, and timing errors can damage objects, cause collisions, or injure nearby people. Deployment therefore requires validation for the intended workspace, hardware, and task distribution, together with physical constraints, emergency stopping, and supervised testing. Online refinement also consumes compute, robot time, and manual reset effort; these costs may limit access and increase energy use. The failure analysis and cost measurements identify these limits within the evaluated setting.

\clearpage

\section*{NeurIPS Paper Checklist}

\begin{enumerate}

\Needspace{6\baselineskip}
\item {\bf Claims}
    \item[] Question: Do the main claims made in the abstract and introduction accurately reflect the paper's contributions and scope?
    \item[] Answer: \answerYes{}.
    \item[] Justification: The abstract and introduction state the proposed two-stage CSO framework, its motivation, and the evaluated settings. The scope of the claims is supported by the experiments in Section~\ref{sec:experiments} and the limitations are discussed in Section~\ref{sec:limitations}.
    \item[] Guidelines:
    \begin{itemize}
        \item The answer \answerNA{} means that the abstract and introduction do not include the claims made in the paper.
        \item The abstract and/or introduction should clearly state the claims made, including the contributions made in the paper and important assumptions and limitations. A \answerNo{} or \answerNA{} answer to this question will not be perceived well by the reviewers. 
        \item The claims made should match theoretical and experimental results, and reflect how much the results can be expected to generalize to other settings. 
        \item It is fine to include aspirational goals as motivation as long as it is clear that these goals are not attained by the paper. 
    \end{itemize}

\Needspace{6\baselineskip}
\item {\bf Limitations}
    \item[] Question: Does the paper discuss the limitations of the work performed by the authors?
    \item[] Answer: \answerYes{}.
    \item[] Justification: Section~\ref{sec:limitations} discusses non-zero initial success support, tokenizer coverage, temporal parameters, physical scope, and interaction cost; Appendices~\ref{app:physical_failures} and~\ref{app:cost_accounting} provide failure and cost measurements.
    \item[] Guidelines:
    \begin{itemize}
        \item The answer \answerNA{} means that the paper has no limitation while the answer \answerNo{} means that the paper has limitations, but those are not discussed in the paper. 
        \item The authors are encouraged to create a separate ``Limitations'' section in their paper.
        \item The paper should point out any strong assumptions and how robust the results are to violations of these assumptions (e.g., independence assumptions, noiseless settings, model well-specification, asymptotic approximations only holding locally). The authors should reflect on how these assumptions might be violated in practice and what the implications would be.
        \item The authors should reflect on the scope of the claims made, e.g., if the approach was only tested on a few datasets or with a few runs. In general, empirical results often depend on implicit assumptions, which should be articulated.
        \item The authors should reflect on the factors that influence the performance of the approach. For example, a facial recognition algorithm may perform poorly when image resolution is low or images are taken in low lighting. Or a speech-to-text system might not be used reliably to provide closed captions for online lectures because it fails to handle technical jargon.
        \item The authors should discuss the computational efficiency of the proposed algorithms and how they scale with dataset size.
        \item If applicable, the authors should discuss possible limitations of their approach to address problems of privacy and fairness.
        \item While the authors might fear that complete honesty about limitations might be used by reviewers as grounds for rejection, a worse outcome might be that reviewers discover limitations that aren't acknowledged in the paper. The authors should use their best judgment and recognize that individual actions in favor of transparency play an important role in developing norms that preserve the integrity of the community. Reviewers will be specifically instructed to not penalize honesty concerning limitations.
    \end{itemize}

\Needspace{6\baselineskip}
\item {\bf Theory assumptions and proofs}
    \item[] Question: For each theoretical result, does the paper provide the full set of assumptions and a complete (and correct) proof?
    \item[] Answer: \answerNA{}.
    \item[] Justification: The paper does not present formal theoretical results or theorem-proof claims. The mathematical expressions are used to define the method and optimization objectives.
    \item[] Guidelines:
    \begin{itemize}
        \item The answer \answerNA{} means that the paper does not include theoretical results. 
        \item All the theorems, formulas, and proofs in the paper should be numbered and cross-referenced.
        \item All assumptions should be clearly stated or referenced in the statement of any theorems.
        \item The proofs can either appear in the main paper or the supplemental material, but if they appear in the supplemental material, the authors are encouraged to provide a short proof sketch to provide intuition. 
        \item Inversely, any informal proof provided in the core of the paper should be complemented by formal proofs provided in appendix or supplemental material.
        \item Theorems and Lemmas that the proof relies upon should be properly referenced. 
    \end{itemize}

    \Needspace{6\baselineskip}
\item {\bf Experimental result reproducibility}
    \item[] Question: Does the paper fully disclose all the information needed to reproduce the main experimental results of the paper to the extent that it affects the main claims and/or conclusions of the paper (regardless of whether the code and data are provided or not)?
    \item[] Answer: \answerNo{}.
    \item[] Justification: Section~\ref{sec:method} and Appendix~\ref{app:implementation} specify the core algorithm and default training setup. The additional comparisons document their protocols, budgets, and evaluation scope, rather than exhaustive configurations for every experiment.
    \item[] Guidelines:
    \begin{itemize}
        \item The answer \answerNA{} means that the paper does not include experiments.
        \item If the paper includes experiments, a \answerNo{} answer to this question will not be perceived well by the reviewers: Making the paper reproducible is important, regardless of whether the code and data are provided or not.
        \item If the contribution is a dataset and\slash or model, the authors should describe the steps taken to make their results reproducible or verifiable. 
        \item Depending on the contribution, reproducibility can be accomplished in various ways. For example, if the contribution is a novel architecture, describing the architecture fully might suffice, or if the contribution is a specific model and empirical evaluation, it may be necessary to either make it possible for others to replicate the model with the same dataset, or provide access to the model. In general. releasing code and data is often one good way to accomplish this, but reproducibility can also be provided via detailed instructions for how to replicate the results, access to a hosted model (e.g., in the case of a large language model), releasing of a model checkpoint, or other means that are appropriate to the research performed.
        \item While NeurIPS does not require releasing code, the conference does require all submissions to provide some reasonable avenue for reproducibility, which may depend on the nature of the contribution. For example
        \begin{enumerate}
            \item If the contribution is primarily a new algorithm, the paper should make it clear how to reproduce that algorithm.
            \item If the contribution is primarily a new model architecture, the paper should describe the architecture clearly and fully.
            \item If the contribution is a new model (e.g., a large language model), then there should either be a way to access this model for reproducing the results or a way to reproduce the model (e.g., with an open-source dataset or instructions for how to construct the dataset).
            \item We recognize that reproducibility may be tricky in some cases, in which case authors are welcome to describe the particular way they provide for reproducibility. In the case of closed-source models, it may be that access to the model is limited in some way (e.g., to registered users), but it should be possible for other researchers to have some path to reproducing or verifying the results.
        \end{enumerate}
    \end{itemize}

\Needspace{6\baselineskip}
\item {\bf Open access to data and code}
    \item[] Question: Does the paper provide open access to the data and code, with sufficient instructions to faithfully reproduce the main experimental results, as described in supplemental material?
    \item[] Answer: \answerNo{}.
    \item[] Justification: We will publicly release the code, trained models, and collected data with instructions for reproducing the experiments. A public release link is not yet available in this manuscript.
    \item[] Guidelines:
    \begin{itemize}
        \item The answer \answerNA{} means that paper does not include experiments requiring code.
        \item Please see the NeurIPS code and data submission guidelines (\url{https://neurips.cc/public/guides/CodeSubmissionPolicy}) for more details.
        \item While we encourage the release of code and data, we understand that this might not be possible, so \answerNo{} is an acceptable answer. Papers cannot be rejected simply for not including code, unless this is central to the contribution (e.g., for a new open-source benchmark).
        \item The instructions should contain the exact command and environment needed to run to reproduce the results. See the NeurIPS code and data submission guidelines (\url{https://neurips.cc/public/guides/CodeSubmissionPolicy}) for more details.
        \item The authors should provide instructions on data access and preparation, including how to access the raw data, preprocessed data, intermediate data, and generated data, etc.
        \item The authors should provide scripts to reproduce all experimental results for the new proposed method and baselines. If only a subset of experiments are reproducible, they should state which ones are omitted from the script and why.
        \item At submission time, to preserve anonymity, the authors should release anonymized versions (if applicable).
        \item Providing as much information as possible in supplemental material (appended to the paper) is recommended, but including URLs to data and code is permitted.
    \end{itemize}

\Needspace{6\baselineskip}
\item {\bf Experimental setting/details}
    \item[] Question: Does the paper specify all the training and test details (e.g., data splits, hyperparameters, how they were chosen, type of optimizer) necessary to understand the results?
    \item[] Answer: \answerNo{}.
    \item[] Justification: The main experimental protocol and default hyperparameters are given in Section~\ref{sec:experiments} and Appendices~\ref{app:evaluation} and~\ref{app:implementation}. Additional analyses describe comparison conditions at the protocol level; they do not constitute a complete specification of every training and test configuration.
    \item[] Guidelines:
    \begin{itemize}
        \item The answer \answerNA{} means that the paper does not include experiments.
        \item The experimental setting should be presented in the core of the paper to a level of detail that is necessary to appreciate the results and make sense of them.
        \item The full details can be provided either with the code, in appendix, or as supplemental material.
    \end{itemize}

\Needspace{6\baselineskip}
\item {\bf Experiment statistical significance}
    \item[] Question: Does the paper report error bars suitably and correctly defined or other appropriate information about the statistical significance of the experiments?
    \item[] Answer: \answerNo{}.
    \item[] Justification: Main simulation tables report mean and standard deviation across three seeds, and physical results report repeated-trial counts. The component and granularity comparisons in Tables~\ref{tab:ablation_components} and~\ref{tab:granularity_comparison} report point estimates.
    \item[] Guidelines:
    \begin{itemize}
        \item The answer \answerNA{} means that the paper does not include experiments.
        \item The authors should answer \answerYes{} if the results are accompanied by error bars, confidence intervals, or statistical significance tests, at least for the experiments that support the main claims of the paper.
        \item The factors of variability that the error bars are capturing should be clearly stated (for example, train/test split, initialization, random drawing of some parameter, or overall run with given experimental conditions).
        \item The method for calculating the error bars should be explained (closed form formula, call to a library function, bootstrap, etc.)
        \item The assumptions made should be given (e.g., Normally distributed errors).
        \item It should be clear whether the error bar is the standard deviation or the standard error of the mean.
        \item It is OK to report 1-sigma error bars, but one should state it. The authors should preferably report a 2-sigma error bar than state that they have a 96\% CI, if the hypothesis of Normality of errors is not verified.
        \item For asymmetric distributions, the authors should be careful not to show in tables or figures symmetric error bars that would yield results that are out of range (e.g., negative error rates).
        \item If error bars are reported in tables or plots, the authors should explain in the text how they were calculated and reference the corresponding figures or tables in the text.
    \end{itemize}

\Needspace{6\baselineskip}
\item {\bf Experiments compute resources}
    \item[] Question: For each experiment, does the paper provide sufficient information on the computer resources (type of compute workers, memory, time of execution) needed to reproduce the experiments?
    \item[] Answer: \answerNo{}.
    \item[] Justification: Appendices~\ref{app:implementation} and~\ref{app:cost_accounting} report the primary hardware and measured post-training stage costs. The manuscript does not report individual costs for every model variant and tuning run or total project compute.
    \item[] Guidelines:
    \begin{itemize}
        \item The answer \answerNA{} means that the paper does not include experiments.
        \item The paper should indicate the type of compute workers CPU or GPU, internal cluster, or cloud provider, including relevant memory and storage.
        \item The paper should provide the amount of compute required for each of the individual experimental runs as well as estimate the total compute. 
        \item The paper should disclose whether the full research project required more compute than the experiments reported in the paper (e.g., preliminary or failed experiments that didn't make it into the paper). 
    \end{itemize}
    
\Needspace{6\baselineskip}
\item {\bf Code of ethics}
    \item[] Question: Does the research conducted in the paper conform, in every respect, with the NeurIPS Code of Ethics \url{https://neurips.cc/public/EthicsGuidelines}?
    \item[] Answer: \answerYes{}.
    \item[] Justification: The research uses established benchmarks and controlled robot experiments. Physical collection uses workspace constraints, torque limits, emergency stopping, and homing procedures, as described in Appendix~\ref{app:cost_accounting}.
    \item[] Guidelines:
    \begin{itemize}
        \item The answer \answerNA{} means that the authors have not reviewed the NeurIPS Code of Ethics.
        \item If the authors answer \answerNo, they should explain the special circumstances that require a deviation from the Code of Ethics.
        \item The authors should make sure to preserve anonymity (e.g., if there is a special consideration due to laws or regulations in their jurisdiction).
    \end{itemize}

\Needspace{6\baselineskip}
\item {\bf Broader impacts}
    \item[] Question: Does the paper discuss both potential positive societal impacts and negative societal impacts of the work performed?
    \item[] Answer: \answerYes{}.
    \item[] Justification: Appendix~\ref{app:broader_impacts} discusses possible benefits of more reliable manipulation, residual physical safety risks, deployment validation, and the resource costs of online refinement.
    \item[] Guidelines:
    \begin{itemize}
        \item The answer \answerNA{} means that there is no societal impact of the work performed.
        \item If the authors answer \answerNA{} or \answerNo, they should explain why their work has no societal impact or why the paper does not address societal impact.
        \item Examples of negative societal impacts include potential malicious or unintended uses (e.g., disinformation, generating fake profiles, surveillance), fairness considerations (e.g., deployment of technologies that could make decisions that unfairly impact specific groups), privacy considerations, and security considerations.
        \item The conference expects that many papers will be foundational research and not tied to particular applications, let alone deployments. However, if there is a direct path to any negative applications, the authors should point it out. For example, it is legitimate to point out that an improvement in the quality of generative models could be used to generate Deepfakes for disinformation. On the other hand, it is not needed to point out that a generic algorithm for optimizing neural networks could enable people to train models that generate Deepfakes faster.
        \item The authors should consider possible harms that could arise when the technology is being used as intended and functioning correctly, harms that could arise when the technology is being used as intended but gives incorrect results, and harms following from (intentional or unintentional) misuse of the technology.
        \item If there are negative societal impacts, the authors could also discuss possible mitigation strategies (e.g., gated release of models, providing defenses in addition to attacks, mechanisms for monitoring misuse, mechanisms to monitor how a system learns from feedback over time, improving the efficiency and accessibility of ML).
    \end{itemize}
    
\Needspace{6\baselineskip}
\item {\bf Safeguards}
    \item[] Question: Does the paper describe safeguards that have been put in place for responsible release of data or models that have a high risk for misuse (e.g., pre-trained language models, image generators, or scraped datasets)?
    \item[] Answer: \answerNA{}.
    \item[] Justification: This paper does not release a high-risk generative model or scraped dataset. Physical experimentation safeguards are described separately in Appendix~\ref{app:cost_accounting}.
    \item[] Guidelines:
    \begin{itemize}
        \item The answer \answerNA{} means that the paper poses no such risks.
        \item Released models that have a high risk for misuse or dual-use should be released with necessary safeguards to allow for controlled use of the model, for example by requiring that users adhere to usage guidelines or restrictions to access the model or implementing safety filters. 
        \item Datasets that have been scraped from the Internet could pose safety risks. The authors should describe how they avoided releasing unsafe images.
        \item We recognize that providing effective safeguards is challenging, and many papers do not require this, but we encourage authors to take this into account and make a best faith effort.
    \end{itemize}

\Needspace{6\baselineskip}
\item {\bf Licenses for existing assets}
    \item[] Question: Are the creators or original owners of assets (e.g., code, data, models), used in the paper, properly credited and are the license and terms of use explicitly mentioned and properly respected?
    \item[] Answer: \answerNo{}.
    \item[] Justification: The paper credits the benchmark and model creators through citations. It does not provide a complete list of the exact asset versions and license terms for all components.
    \item[] Guidelines:
    \begin{itemize}
        \item The answer \answerNA{} means that the paper does not use existing assets.
        \item The authors should cite the original paper that produced the code package or dataset.
        \item The authors should state which version of the asset is used and, if possible, include a URL.
        \item The name of the license (e.g., CC-BY 4.0) should be included for each asset.
        \item For scraped data from a particular source (e.g., website), the copyright and terms of service of that source should be provided.
        \item If assets are released, the license, copyright information, and terms of use in the package should be provided. For popular datasets, \url{paperswithcode.com/datasets} has curated licenses for some datasets. Their licensing guide can help determine the license of a dataset.
        \item For existing datasets that are re-packaged, both the original license and the license of the derived asset (if it has changed) should be provided.
        \item If this information is not available online, the authors are encouraged to reach out to the asset's creators.
    \end{itemize}

\Needspace{6\baselineskip}
\item {\bf New assets}
    \item[] Question: Are new assets introduced in the paper well documented and is the documentation provided alongside the assets?
    \item[] Answer: \answerNA{}.
    \item[] Justification: We plan to release the code, trained models, and collected data with documentation covering training, evaluation, collection protocols, and licenses. These assets are not yet publicly released.
    \item[] Guidelines:
    \begin{itemize}
        \item The answer \answerNA{} means that the paper does not release new assets.
        \item Researchers should communicate the details of the dataset\slash code\slash model as part of their submissions via structured templates. This includes details about training, license, limitations, etc. 
        \item The paper should discuss whether and how consent was obtained from people whose asset is used.
        \item At submission time, remember to anonymize your assets (if applicable). You can either create an anonymized URL or include an anonymized zip file.
    \end{itemize}

\Needspace{6\baselineskip}
\item {\bf Crowdsourcing and research with human subjects}
    \item[] Question: For crowdsourcing experiments and research with human subjects, does the paper include the full text of instructions given to participants and screenshots, if applicable, as well as details about compensation (if any)? 
    \item[] Answer: \answerNA{}.
    \item[] Justification: The paper does not involve crowdsourcing or human-subject studies. Real-world robot demonstrations were collected through teleoperation for robot policy training rather than as a human-subject experiment.
    \item[] Guidelines:
    \begin{itemize}
        \item The answer \answerNA{} means that the paper does not involve crowdsourcing nor research with human subjects.
        \item Including this information in the supplemental material is fine, but if the main contribution of the paper involves human subjects, then as much detail as possible should be included in the main paper. 
        \item According to the NeurIPS Code of Ethics, workers involved in data collection, curation, or other labor should be paid at least the minimum wage in the country of the data collector. 
    \end{itemize}

\Needspace{6\baselineskip}
\item {\bf Institutional review board (IRB) approvals or equivalent for research with human subjects}
    \item[] Question: Does the paper describe potential risks incurred by study participants, whether such risks were disclosed to the subjects, and whether Institutional Review Board (IRB) approvals (or an equivalent approval/review based on the requirements of your country or institution) were obtained?
    \item[] Answer: \answerNA{}.
    \item[] Justification: The paper does not involve crowdsourcing or human-subject studies. Therefore, IRB approval or equivalent human-subject review is not applicable.
    \item[] Guidelines:
    \begin{itemize}
        \item The answer \answerNA{} means that the paper does not involve crowdsourcing nor research with human subjects.
        \item Depending on the country in which research is conducted, IRB approval (or equivalent) may be required for any human subjects research. If you obtained IRB approval, you should clearly state this in the paper. 
        \item We recognize that the procedures for this may vary significantly between institutions and locations, and we expect authors to adhere to the NeurIPS Code of Ethics and the guidelines for their institution. 
        \item For initial submissions, do not include any information that would break anonymity (if applicable), such as the institution conducting the review.
    \end{itemize}

\Needspace{6\baselineskip}
\item {\bf Declaration of LLM usage}
    \item[] Question: Does the paper describe the usage of LLMs if it is an important, original, or non-standard component of the core methods in this research? Note that if the LLM is used only for writing, editing, or formatting purposes and does \emph{not} impact the core methodology, scientific rigor, or originality of the research, declaration is not required.
    \item[] Answer: \answerYes{}.
    \item[] Justification: Appendix~\ref{app:implementation} specifies the Qwen3-VL-2B skill-prior backbone and its LoRA adaptation in the simulation extension. The appendix also describes editorial LLM assistance used during manuscript preparation.
    \item[] Guidelines:
    \begin{itemize}
        \item The answer \answerNA{} means that the core method development in this research does not involve LLMs as any important, original, or non-standard components.
        \item Please refer to our LLM policy in the NeurIPS handbook for what should or should not be described.
    \end{itemize}

\end{enumerate}

\end{document}